\documentclass[a4paper,fleqn]{cas-dc}
\usepackage[authoryear,round]{natbib}
\usepackage{textcomp}
\usepackage{tikz}
\usepackage{forest}
\usepackage{xcolor}
\usepackage{caption}
\usepackage{float}
\usepackage{amsmath,amssymb,amsfonts}
\usepackage{nccmath}
\usepackage{multirow}
\usepackage{dsfont} 
\usepackage{placeins}
\usepackage{booktabs,tabularx,makecell,array}
\usepackage{amssymb} 
\usepackage{stfloats}
\usepackage{pifont}
\restylefloat{figure}
\def\tsc#1{\csdef{#1}{\textsc{\lowercase{#1}}\xspace}}
\tsc{WGM}
\tsc{QE}
\tsc{EP}
\tsc{PMS}
\tsc{BEC}
\tsc{DE}
\begin{document}
\let\WriteBookmarks\relax
\def\floatpagepagefraction{1}
\def\textpagefraction{.001}
\shortauthors{S. Zhou et~al.}
\shorttitle{A Review of Longitudinal Radiology Report Generation}
\title[mode=title]{A Review of Longitudinal Radiology Report Generation: Dataset Composition, Methods, and Performance Evaluation}

\author[1]{Shaoyang Zhou}
\fnmark[1]
\ead{szho0189@uni.sydney.edu.au}
\author[1]{Yingshu Li}
\fnmark[1]
\ead{yingshu.li@sydney.edu.au}

\author[1]{Yunyi Liu}
\fnmark[1]
\ead{yunyi.liu1@sydney.edu.au}

\author[2]{Lingqiao Liu}
\ead{lingqiao.liu@adelaide.edu.au}

\author[3]{Lei Wang}
\ead{leiw@uow.edu.au}

\author[1]{Luping Zhou}[orcid=0000-0001-8762-2424]
\cormark[1]
\ead{luping.zhou@sydney.edu.au}

\affiliation[1]{organization={School of Electrical and Computer Engineering, The University of Sydney},
                city={Sydney},
                state={NSW},
                postcode={2006},
                country={Australia}}

\affiliation[2]{organization={School of Computer Science, The University of Adelaide},
                city={Adelaide},
                state={SA},
                postcode={5005},
                country={Australia}}

\affiliation[3]{organization={School of Computing and Information Technology, University of Wollongong},
                city={Wollongong},
                state={NSW},
                postcode={2522},
                country={Australia}}

\cortext[1]{Corresponding author.}
\fntext[1]{Co-first authors.}

\begin{abstract}
Chest X-ray imaging is a widely used diagnostic tool in modern medicine, and its high utilization creates substantial workloads for radiologists. To alleviate this burden, vision-language models are increasingly applied to automate Chest X-ray radiology report generation (CXR-RRG), aiming for clinically accurate descriptions while reducing manual effort. Conventional approaches, however, typically rely on single-image, failing to capture the longitudinal context necessary for producing clinically faithful comparison statements. Recently, growing attention has been directed toward incorporating longitudinal data into CXR-RRG, enabling models to leverage historical studies in ways that mirror radiologists’ diagnostic workflows. Nevertheless, existing surveys primarily address single-image CXR-RRG and offer limited guidance for longitudinal settings, leaving researchers without a systematic framework for model design. To address this gap, this survey provides the first comprehensive review of longitudinal radiology report generation (LRRG). Specifically, we examine dataset construction strategies, report generation architectures alongside longitudinally tailored designs, and evaluation protocols encompassing both longitudinal-specific measures and widely used benchmarks. We further summarize LRRG methods' performance, alongside analyses of different ablation studies, which collectively highlight the critical role of longitudinal information and architectural design choices in improving model performance. Finally, we summarize five major limitations of current research and outline promising directions for future development, aiming to lay a foundation for advancing this emerging field.
\end{abstract}




\maketitle

\section{Introduction}
\label{sec:introduction}
Radiology reports are structured medical documents that synthesize imaging observations, clinical context, and diagnostic impression, serving as a critical communication bridge between radiologists and referring physicians. In recent years, the workload of radiologists has increased substantially~\citep{cao2023current,bailey2022understanding}, driven by the growing demand for imaging studies and the complexity of diagnostic tasks. This is particularly evident in Chest X-ray examinations, which are among the most widely used imaging modalities, making Chest X-ray radiology reports a key component of clinical diagnosis. Addressing this growing workload underscores the urgent need for solutions that can improve reporting efficiency while maintaining high diagnostic quality.

\begin{figure}[t]
    \includegraphics[width=\columnwidth,height=0.35\textheight,keepaspectratio]{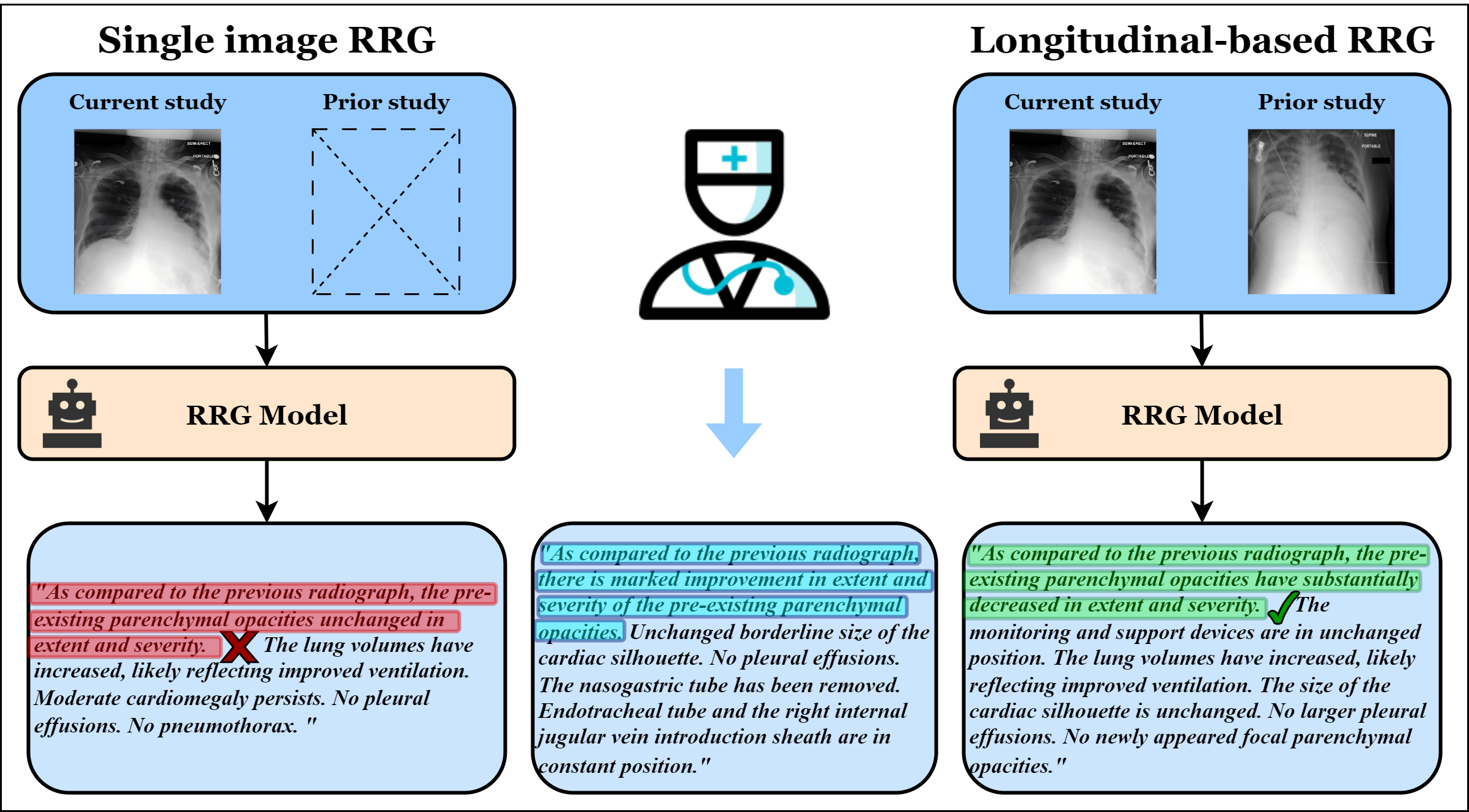}
    \caption{Examples illustrating the difference between single-image and LRRG settings. Single-image RRG: Without prior studies, the model fails to provide accurate comparison statements with prior examinations. LRRG: Access to prior studies enables the model to generate comparison statements closely aligned with those in the ground-truth report.}
    \label{fig:introduction}
\end{figure}

Among the emerging solutions, automated radiology report generation (RRG)~\citep{rennie2017self} has shown promise as an effective strategy to alleviate radiologists’ workload~\citep{najdenkoska2022uncertainty}. By assisting in the generation of accurate and clinically meaningful free-text reports, current RRG systems aim to meet the demand for fine-grained descriptions and precise diagnostic terminology in Chest X-ray reports, and they have achieved notable progress. However, most of these methods rely solely on a single Chest X-ray image~\citep{lee2025cxr,wang2023r2gengpt,li2024kargen}, overlooking the rich temporal information available in clinical datasets. For instance, in the MIMIC-CXR dataset, more than 67\% of patients undergo multiple examinations, and such temporal information is routinely leveraged in radiologists’ diagnostic workflows to improve accuracy~\citep{kelly2012chest}. Without access to prior studies, the model may fail to capture temporal progression in the report, as illustrated in Fig.~\ref{fig:introduction}. Prior work~\citep{hyland2023maira} further showed that the absence of such information often causes models to hallucinate spurious prior references. Collectively, these observations underscore that integrating longitudinal information into model architectures is not merely advantageous but essential for producing reliable and clinically faithful CXR-RRG.

Building on this insight, recent CXR-RRG studies~\citep{song2025ddatr,hou2023recap,liu2025enhanced,zhang2024libra,hou2025radar,miao2024evoke,bannur2024maira,bannur2023learning,park2024m4cxr,yang2025spatio,zhou2024medversa,wang2024hergen,serra2023controllable,sanjeev2024tibix,wang2025llm,nicolson2024longitudinal,zhu2023utilizing,santiesteban2024enhancing,liu2025hc} have increasingly incorporated longitudinal Chest X-ray data to enhance temporal reasoning and improve clinical accuracy. For example, some methods~\citep{nicolson2024longitudinal,bannur2024maira,bannur2023learning,sanjeev2024tibix,park2024m4cxr,wang2024hergen,zhu2023utilizing, liu2025enhanced, zhang2024libra, wang2025llm, santiesteban2024enhancing} leverage attention mechanisms to fuse features across time, enabling the model to better align prior information with current report generation, while others~\citep{liu2025hc,hou2023recap,wang2024hergen, bannur2023learning,sanjeev2024tibix} introduce temporal change classification or explicitly encode the interval between examinations to capture disease progression. Collectively, these approaches not only highlight the promise of longitudinal modeling in RRG but also establish it as a rapidly expanding research frontier.

Generally speaking, prior surveys~\citep{sloan2024automated,mamdouh2025advancements,min2025automating} have provided broad overviews of RRG methods, but they mainly focus on single-image-based approaches and overlook the unique challenges of longitudinal settings. With the growing recognition of the potential of longitudinal data and the rapid emergence of diverse research issues, methodological directions, architectures, and dataset designs, there is a timely need for a dedicated survey on the LRRG task. To address this gap, the present work focuses exclusively on longitudinal CXR-RRG approaches. It is worth noting that although this survey centers on Chest X-ray images, the modeling principles discussed herein are inherently modality-agnostic and could be extended to other imaging domains, such as CT, MRI, and PET. We systematically categorize existing methods according to how they exploit temporal information and highlight promising avenues for future research. The main contributions of this survey are summarized as follows:

\begin{itemize}
\item We survey recent advances in LRRG, covering dataset construction strategies, model architectures for report generation, longitudinal integration challenges and approaches, and evaluation protocols, including longitudinal-specific metrics.
\item We aggregate published results on representative LRRG methods and collate reported ablations to elucidate the effects of integration strategies and architectural choices.
\item We summarize current limitations, and offer practical recommendations and future directions for incorporating longitudinal data into CXR-RRG systems.
\end{itemize}

The remainder of this survey is organized as follows: Section~\ref{sec:Datasets} reviews different methods for constructing longitudinal datasets from MIMIC-CXR. Section~\ref{sec:Radiology report generation} reviews architectural designs for radiology report generation, highlighting the configurations of vision encoders and text decoders across different longitudinal methods. Section~\ref{sec:Core Challenges and Approaches in Longitudinal Report Generation} categorizes approaches that leverage longitudinal data to enhance model performance. Section~\ref{sec:Evaluation Metrics} presents a range of commonly used evaluation metrics, along with longitudinal-specific measures. Section~\ref{sec:Discussion} emphasizes the critical role of longitudinal data and the contribution of distinct structural components, while also consolidating the reported performance of existing LRRG methods. Finally, Section~\ref{sec:Current limitations and future improvements} outlines current limitations and potential directions for future improvements.

  \begin{figure*}[t!]
    \centering
    \includegraphics[width=\linewidth]{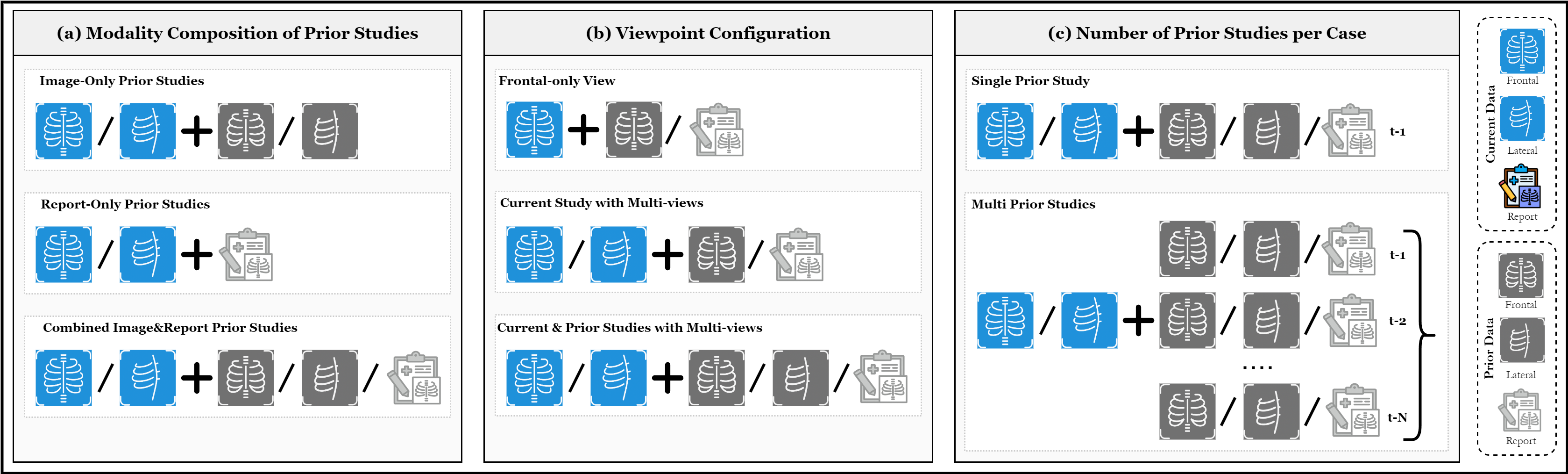}
    \caption{Dataset construction strategies for the LRRG task, which can be divided into three categories: (a) Modality Composition of Prior Studies: prior studies may contain only images, only reports, or both. (b) Viewpoint Configuration: studies may consider only frontal views, multiple views for the current study, or multiple views for both current and prior studies. (c) Longitudinal dataset scale – Number of prior studies per case: each case may include either a single prior study or multiple temporally ordered prior studies. }
    \label{fig:Examples of dataset}
\end{figure*}

\begin{figure}[!h]
    \centering    \includegraphics[width=\columnwidth]{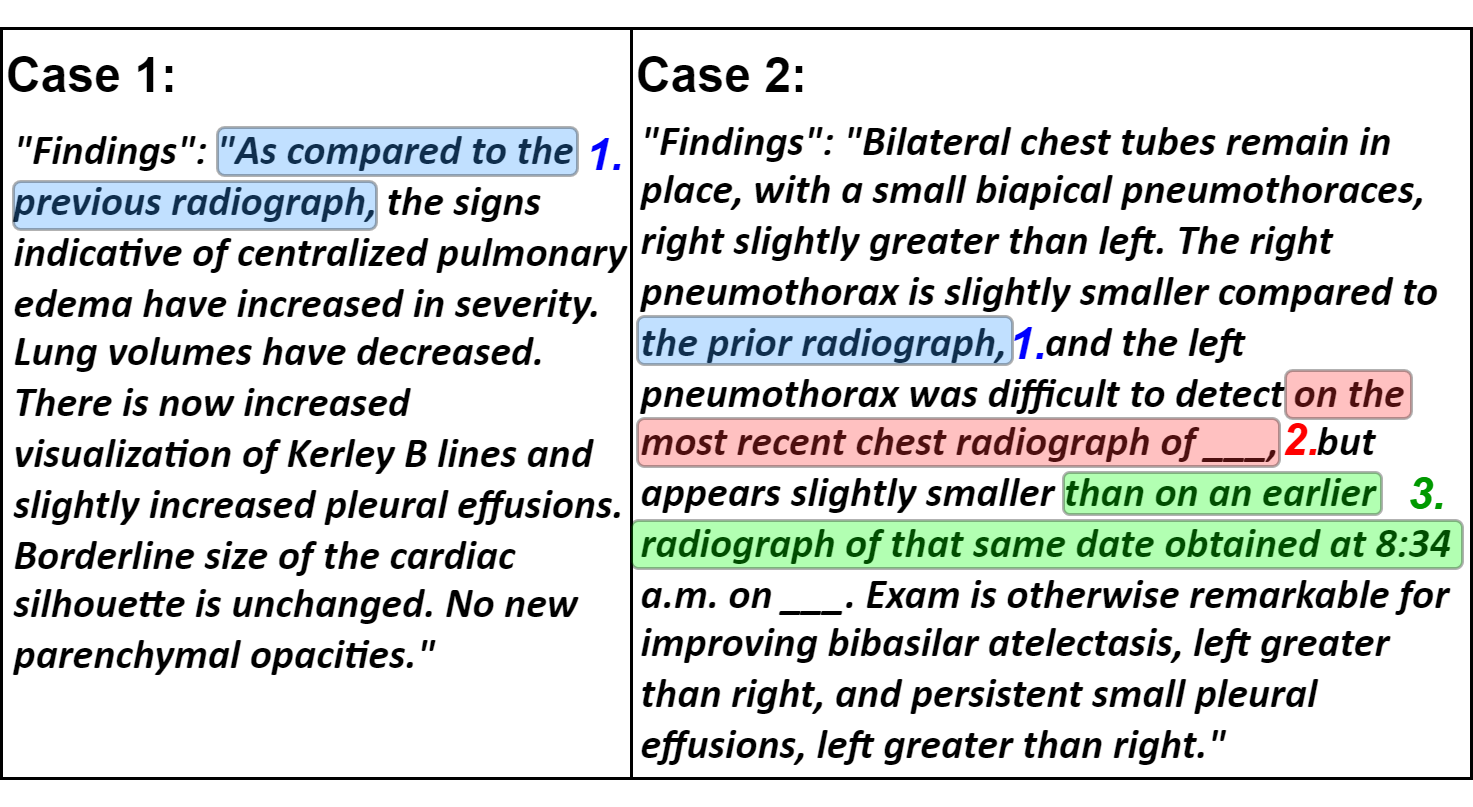}
    \caption{Examples from the MIMIC-CXR dataset demonstrate varying numbers of prior comparisons of different reports. Case 1 illustrates a single-time-point comparison, where findings are referenced only against the most recent prior study. Case 2 illustrates a multi-time-point comparison, where findings are evaluated against multiple prior studies, including the most recent, recent, and earlier dated radiographs.}
    \label{fig:single and multiple}
\end{figure}

\section{Datasets}
\label{sec:Datasets}
Effective utilization of longitudinal information in the RRG task depends critically on the availability, composition, and organization of prior studies within the dataset. Recent works have adopted diverse strategies for constructing longitudinal datasets, as illustrated in Fig.~\ref{fig:Examples of dataset}. They focus on the MIMIC-CXR dataset~\citep{johnson2019mimic}, which contains 473,057 images and 227,835 reports from 63,478 patients. Among these data, the training set includes 270,790 samples, the validation set 2,130, and the test set 3,858. Within this dataset, studies for each patient are commonly ordered by the keys Study-Date and Study-Time to establish the temporal sequence. To introduce different methods in detail, we: (i) outline prior-modality choices; (ii) describe dataset scale factors; and (iii) review viewpoint use in the following subsections.

\subsection{Modality Composition of Prior Studies}
In the traditional RRG task, each study typically comprises two modalities: Chest X-ray images and the corresponding reference report, representing the patient’s current examination. In the longitudinal setting, however, the current study may also incorporate prior studies' data, which can include various combinations of historical modalities. Based on how these prior modalities are composed, existing approaches can be broadly categorized into three groups:

\subsubsection{Image-Only Prior Studies}
These methods~\citep{zhou2024medversa,zhang2024libra,wang2024hergen,serra2023controllable,santiesteban2024enhancing,sanjeev2024tibix} include only the prior image as longitudinal input.
The main purpose of introducing the prior image is to enable cross-temporal fusion of visual features, thereby enhancing visual representations and improving alignment with the generated report.

\subsubsection{Report-Only Prior Studies}
Similar to the single-modality setting, some models~\citep{wang2025llm,nicolson2024longitudinal} rely solely on the prior report.
The prior report is typically used as a textual prompt to enrich cross-temporal contextual information, guiding the model toward clinically coherent report generation.

\subsubsection{Combined Image \& Report Prior Studies}
Different from the above single-modality approaches, many recent works~\citep{song2025ddatr,hou2023recap,zhu2023utilizing,liu2025enhanced,hou2025radar,miao2024evoke,bannur2024maira,bannur2023learning,liu2025hc,park2024m4cxr,yang2025spatio,huang2024hist} leverage both prior images and reports.
This multi-modal combination enables the model to jointly capture complementary temporal cues from visual and textual modalities, thereby supporting richer temporal reasoning and improving alignment of disease progression across modalities.
\subsection{Longitudinal Dataset Scale}
Longitudinal data across modalities provide richer temporal representations. However, since not all patients undergo prior examinations, datasets are typically organized under two settings: including or excluding non-longitudinal cases. When restricted to cases with longitudinal data and processed following the \textit{ULCX}~\citep{zhu2023utilizing} strategy, the longitudinal MIMIC-CXR subset, distinct from the original dataset with non-longitudinal cases, comprises 95,168 samples from 26,625 patients (92,374 for training, 737 for validation, and 2,058 for testing).

Beyond the inclusion of non-longitudinal cases, another key factor influencing the scale of a longitudinal dataset is the number of prior studies available per case. Consequently, existing methodologies can be categorized by the number of prior studies they incorporate for each patient.

\subsubsection{Single Prior Study}
As emphasized by \textit{HIST-AID}~\citep{huang2024hist}, the most recent examination of the patient often improves model performance, while older examinations may introduce misleading information. Therefore, most models~\citep{song2025ddatr, hou2023recap, zhang2024libra, hou2025radar, zhu2023utilizing, nicolson2024longitudinal, bannur2023learning, serra2023controllable, wang2025llm, sanjeev2024tibix, liu2025hc, bannur2024maira, park2024m4cxr, zhou2024medversa} that utilize a single prior study only consider the most recent one. This strategy also helps reduce input token length and computational cost.

\subsubsection{Multiple Recent Prior Studies}
To capture a more comprehensive trajectory of disease progression and accommodate cases where reports compare multiple time points (see Fig.~\ref{fig:single and multiple}), several models~\citep{liu2025enhanced, miao2024evoke, wang2024hergen, yang2025spatio, huang2024hist} incorporate information from multiple prior studies. For example, \textit{MLRG}~\citep{liu2025enhanced}, \textit{EVOKE}~\citep{miao2024evoke}, and \textit{STREAM}~\citep{yang2025spatio} utilize both the most recent and the second most recent studies, whereas \textit{HERGen}~\citep{wang2024hergen} leverages up to five prior studies per patient. \textit{HIST-AID}~\citep{huang2024hist} is particularly notable, extending the temporal horizon further by integrating an average of 11 reports and 13 images per patient across 12,221 patients, thereby enabling more comprehensive modeling of long-term disease progression.

\subsection{Multi-Viewpoint Utilization}
While longitudinal data primarily determine the depth of a dataset, its width is governed by the availability of the frontal and lateral viewpoints. In particular, the lateral Chest X-ray not only complements the frontal view but can also serve as a primary source of visual information, thereby enhancing model robustness and overall performance. Compared with the original MIMIC-CXR dataset, restricting the dataset to frontal views only reduces its scale to 166,702 samples in total, with 162,955 for training, 1,286 for validation, and 2,461 for testing, as configured in \textit{Libra}~\citep{zhang2024libra}. To further illustrate how these two viewpoints are leveraged, existing approaches can be broadly grouped into three categories:

\subsubsection{Multi-View Representation of the Current Study}
To better enhance the current time point report generation, some models~\citep{hou2023recap,bannur2024maira, wang2025llm,nicolson2024longitudinal} exploit both frontal and lateral viewpoints of the current study to enhance its visual feature representation.

\subsubsection{Multi-View Representation of Current and Prior Studies}
In longitudinal datasets, certain records may lack a frontal view while retaining the corresponding lateral view. To handle such incomplete cases, several models~\citep{song2025ddatr,zhu2023utilizing,liu2025enhanced,miao2024evoke,zhou2024medversa,liu2025hc,yang2025spatio,park2024m4cxr} exploit the available lateral images from prior studies to supplement the visual representation of previous time points. When both views are available, the lateral view serves as an additional source of contextual information, further enriching the visual representation of prior studies.

\subsubsection{Single-View Representation (Frontal Only)}
Focusing exclusively on longitudinal information, other models~\citep{zhang2024libra,hou2025radar,wang2024hergen,bannur2023learning,serra2023controllable,sanjeev2024tibix,santiesteban2024enhancing} adopt only the frontal viewpoint, thereby circumventing the increased complexity associated with multi-view feature fusion.


\definecolor{TMIBlue}{RGB}{25,65,145}
\definecolor{TMIBlueLight}{RGB}{100,130,200}
\definecolor{TMIBluePale}{RGB}{220,230,250}

\tikzstyle{base-node}=[
    rectangle,
    rounded corners=4pt,
    text opacity=1,
    inner sep=4pt,
    align=center,
    fill opacity=0.95,
    draw opacity=0.8,
]

\tikzstyle{root_node}=[
    base-node,
    fill=TMIBlue,
    text=white,
    font=\small\bfseries,
    minimum height=3.2em,
    text width=18em,
    draw=TMIBlue!90!black,
    line width=1pt,
]

\tikzstyle{level1_node}=[
    base-node,
    fill=TMIBlue!75,
    text=white,
    font=\footnotesize\bfseries,
    minimum height=2.2em,
    text width=8em,
    draw=TMIBlue!85,
    line width=0.8pt,
]

\tikzstyle{level2_node}=[
    base-node,
    fill=TMIBlue!50,
    text=black,
    font=\scriptsize\bfseries,
    minimum height=2em,
    text width=8em,
    draw=TMIBlue!65,
    line width=0.6pt,
]

\tikzstyle{leaf_node}=[
    base-node,
    fill=TMIBluePale,
    text=black,
    font=\scriptsize,
    minimum height=1.8em,
    text width=15em,
    align=left,
    inner xsep=6pt,
    inner ysep=4pt,
    draw=TMIBlueLight!60,
    line width=0.5pt,
]

\begin{figure*}[htbp]
    \centering
    \resizebox{0.98\textwidth}{!}{
        \begin{forest}
            for tree={
                grow=east,
                reversed=true,
                anchor=base west,
                parent anchor=east,
                child anchor=west,
                base=left,
                edge={TMIBlue!70, line width=1.5pt, rounded corners=2pt},
                edge path={
                    \noexpand\path [draw, \forestoption{edge}] 
                    (!u.parent anchor) -- ++(10pt,0) |- (.child anchor)\forestoption{edge label};
                },
                s sep=6pt,
                l sep=18pt,
                ver/.style={rotate=90, child anchor=north, parent anchor=south, anchor=center},
            },
            [
                {Core Challenges and Approaches in\\the LRRG task.}, 
                root_node, ver
                [
                    {Handling Missing\\Prior Data}, level1_node
                    [Zero Padding for Missing Prior Studies~\citep{serra2023controllable,wang2025llm,sanjeev2024tibix,huang2024hist,song2025ddatr}, leaf_node]
                    [Special Token-Based Representation for Absent Inputs~\citep{liu2025enhanced,bannur2023learning}, leaf_node]
                    [Prior Image Imputation via Current Image Duplication~\citep{zhang2024libra,hou2025radar}, leaf_node]
                    [Using Pseudo Prior Reports for Prompting~\citep{nicolson2024longitudinal}, leaf_node]
                ]
                [
                    {Cross-Time and\\Cross-Modal\\Feature Alignment}, level1_node
                    [Contrastive Learning Foundations~\citep{radford2021learning}, leaf_node]
                    [Batch-Level Alignment across Studies~\citep{liu2025enhanced,bannur2023learning}, leaf_node]
                    [Patient-Level Alignment across Time~\citep{wang2024hergen,liu2025hc}, leaf_node]
                ]
                [
                    {Longitudinal Feature\\Fusion Strategies}, level1_node
                    [Concatenation-Based Methods~\citep{serra2023controllable,wang2024hergen}, leaf_node] ,
                    [Gated Mechanism-Based Methods~\citep{hou2023recap,song2025ddatr} , leaf_node]                    
                    [
                        {Attention-Based Fusion}, level2_node
                        [Token-Level Self-Attention Fusion~\citep{nicolson2024longitudinal,bannur2024maira,bannur2023learning,sanjeev2024tibix,park2024m4cxr,huang2024hist,wang2024hergen}, leaf_node]
                        [Query-Based Cross-Attention Fusion~\citep{zhu2023utilizing,liu2025enhanced,zhang2024libra,wang2025llm,santiesteban2024enhancing}
, leaf_node]
                    ]
                ]
                [
                    {Auxiliary Enhancement\\Modules}, level1_node
                    [Classification of Temporal Changes~\citep{liu2025hc,hou2023recap}, leaf_node]
                    [Retrieval-Based Augmentation~\citep{hou2023recap,yang2025spatio}, leaf_node]
                    [Time Gap Encoding~\citep{wang2024hergen,bannur2023learning,sanjeev2024tibix,huang2024hist} , leaf_node]
                ]
            ]
        \end{forest}
    }
    \caption{
        Hierarchical taxonomy of core challenges and technical strategies for longitudinal medical report generation. 
    }
    \label{fig:longitudinal_taxonomy}
\end{figure*}
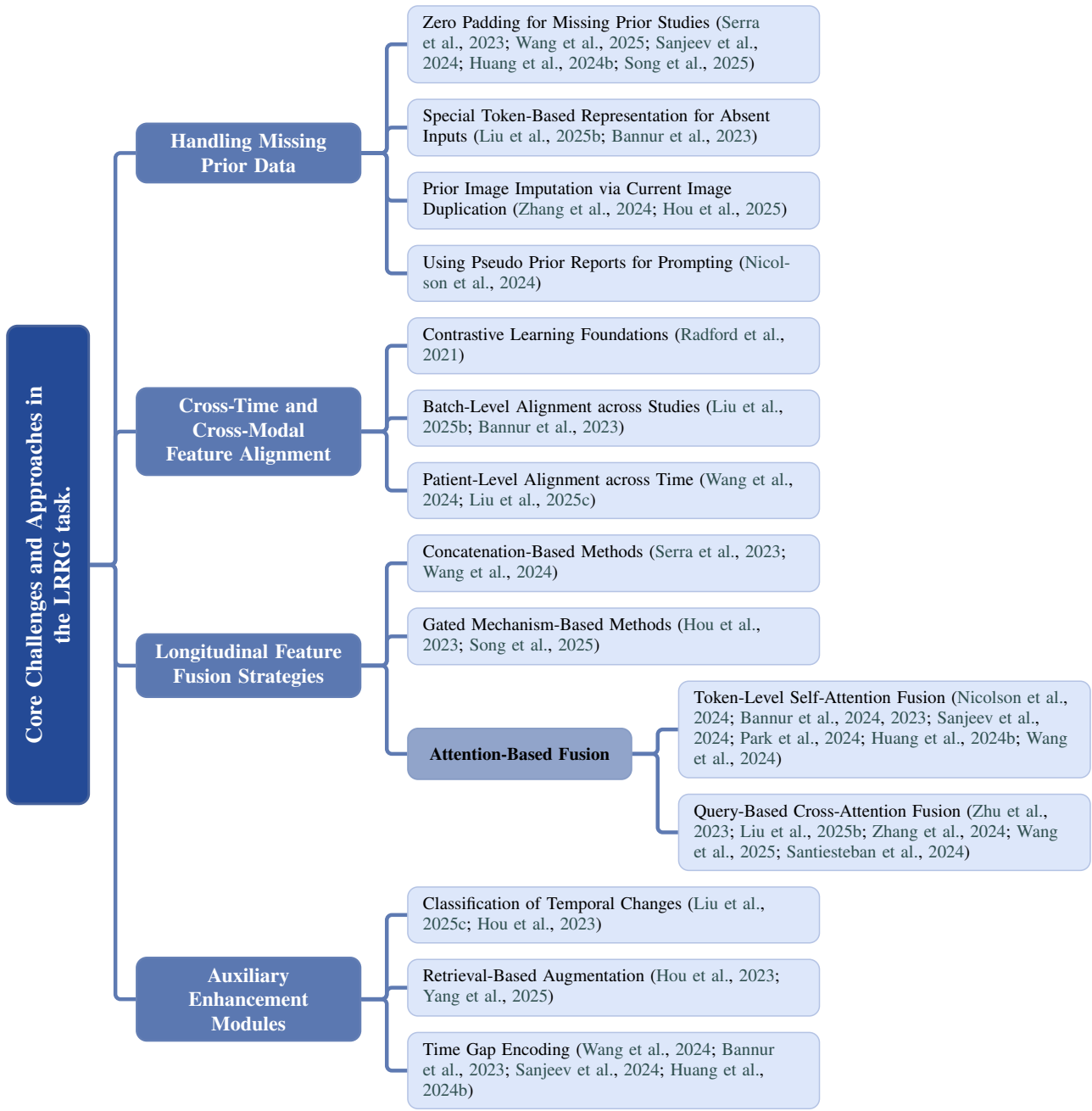

\section{Radiology Report Generation}
\label{sec:Radiology report generation}

In the RRG task, model architectures are generally structured around three core components: vision encoder, text decoder, and loss function. The vision encoder extracts high-dimensional visual representations from Chest X-ray images. These representations, along with auxiliary prompts, serve as conditioning signals for the text decoder, which generates coherent and clinically accurate radiology reports. Finally, the loss function typically guides the model to maximize the conditional probability of each report token given the preceding context. This section systematically reviews how recent longitudinal methods configure and adapt these three components.

\subsection{Visual Encoder}
A robust vision encoder is pivotal for deriving semantically rich and spatially precise representations from Chest X-ray images. In longitudinal settings, it must not only capture fine-grained anatomical details but also enable consistent comparison and fusion of features across different time points. To achieve this, recent methods employ high-capacity CNNs, Transformer-based architectures, or region-level feature extractors, thereby generating high-quality visual prompts to effectively guide the text decoder.
\subsubsection{Vision Backbones for Holistic Feature Extraction}
Existing LRRG models typically employ vision encoders to process the current and prior images separately, yielding a unified sequence of visual tokens that is subsequently consumed by downstream modules, as illustrated in Section~\ref{sec:Core Challenges and Approaches in Longitudinal Report Generation}. Regarding the detailed vision encoder configurations, several approaches~\citep{zhu2023utilizing,bannur2023learning} employ convolutional neural networks (CNNs) to better capture local visual patterns.
Given the strong spatial modeling capabilities of Transformer architectures, many recent models prefer purely Transformer-based encoders. For instance, \textit{RAD-DINO}~\citep{perez2024rad} is commonly employed for high-resolution images (518$\times$518) due to its fine-grained spatial understanding in some models~\citep{zhang2024libra,wang2025llm,bannur2024maira}, while other models~\citep{yang2025spatio,liu2025hc} use \textit{Swin Transformers}~\citep{liu2021swin} for low-resolution inputs (224$\times$224), incorporating local attention mechanisms to enhance the model’s perception of regional visual patterns.

\subsubsection{Region-Based Feature Extraction}
Despite employing powerful vision encoders, several models~\citep{bannur2024maira,serra2023controllable,yang2025spatio,santiesteban2024enhancing} incorporate object detection modules into their architectures. Most of these methods employ \textit{Faster R-CNN}~\citep{ren2016faster} to extract region-of-interest (ROI) features from input images for downstream tasks. \textit{MAIRA-2}~\citep{bannur2024maira} uses it to detect ROI positions, which are incorporated into the structured label of the report to highlight the spatial regions attended to during generation. Similarly, \textit{ERRG}~\citep{santiesteban2024enhancing} utilizes the bounding boxes' positions as part of the report prompt, allowing the model to ground its generation in specific anatomical regions. In contrast, some approaches integrate detection to better capture longitudinal feature representations. \textit{CCXRG}~\citep{serra2023controllable} employs Faster R-CNN as the image encoder and exclusively uses the extracted ROI-level visual features from both the current and prior images, focusing on localized region-based representation.

\textit{STREAM}~\citep{yang2025spatio} performs a more integrated design at the encoder stage: it first extracts ROI regions, and then matches them against an auxiliary ROI library, which contains descriptions and temporal metadata of similar historical regions. This enables the model to incorporate prior knowledge and region-level longitudinal context into the generation process.

\subsection{Text Decoder}
After obtaining the visual prompts $Pro_{vision}$, these feature representations are integrated with auxiliary prompts to guide the text decoder. Such auxiliary cues play a critical role in directing the model’s attention toward salient findings. Consequently, many existing approaches focus on enriching or refining these prompts to enhance the accuracy, coherence, and clinical relevance of the generated reports.

In the text decoder, the probability of generating the report sequence \( R = \{r_1, r_2, \dots, r_L\} \) is typically modeled as:
\begin{equation}
p(R) \sim \prod_{t=1}^{L} p_{\theta}\bigl(r_t \mid X\bigr),
\label{eq:report_gen_prob}
\end{equation}
where \( L \) is the report length, X represents the prompt information.

In the conventional single-image setting, prior works have shown that incorporating structured report components—such as Indication, Comparison, History, and Technique—as the clinical prompts $Pro_{clinical}$ can significantly improve generation quality. Accordingly, the prompt set X can be expanded as:
\begin{equation}
X \in \{\mathrm{Pro}_{\mathrm{vision}},\ \mathrm{Pro}_{\mathrm{clinical}},\ S_{sys},\ R_{0:t-1}\}.
\label{eq:report_gen_input_single}
\end{equation}
\( S_{sys} \) denotes the system-level directive (e.g., “You are an assistant in radiology, responsible for analyzing medical imaging studies and generating detailed, structured, and accurate radiology reports.”), and \( R_{0:t-1} \) is the partially generated report sequence.

In the longitudinal setting, several models~\citep{hou2025radar,nicolson2024longitudinal,bannur2024maira,bannur2023learning} extend the input by incorporating the prior report as an additional prompt, \( \mathrm{Pro}_{\mathrm{prior-findings}} \). The input configuration becomes:

\begin{equation}
X \in \{\mathrm{Pro}_{\mathrm{vision}},\ \mathrm{Pro}_{\mathrm{clinical}},\ 
\mathrm{Pro}_{\mathrm{prior-findings}},\ S_{sys},\ R_{0:t-1}\},
\label{eq:report_gen_input_long}
\end{equation}
allowing the model to maintain temporal continuity and align current observations with historical findings.

In addition to textual prompts, \textit{ERRG}~\citep{santiesteban2024enhancing} introduces a spatial prompting mechanism \( \mathrm{Pro}_{\mathrm{spatial}} \), where the image is divided into \( P \times P \) patches and bounding box coordinates are quantized to the centers of these patches. These spatial prompts help direct attention to regions of interest:
\begin{equation}
X \in \{\mathrm{Pro}_{\mathrm{vision}},\ \mathrm{Pro}_{\mathrm{spatial}},\ S_{sys},\ R_{0:t-1}\}.
\label{eq:report_gen_input_spatial}
\end{equation}

More recently, \textit{STREAM}~\citep{yang2025spatio} proposes a retrieval-augmented strategy that incorporates region-level textual descriptions as prompts \( \mathrm{Pro}_{\mathrm{text}} \). The model retrieves the top-\( k \) most similar regions from a memory bank. For each candidate, if its prior region is sufficiently similar to the input’s prior region (based on a threshold), its textual description is used to update the prompt:
\begin{equation}
X \in \{\mathrm{Pro}_{\mathrm{vision}},\ \mathrm{Pro}_{\mathrm{text}},\ S_{sys},\ R_{0:t-1}\}.
\label{eq:report_gen_input_retrieval}
\end{equation}
This retrieval-based prompting introduces fine-grained spatial and temporal cues, enhancing the contextual fidelity and clinical relevance of longitudinal radiology report generation.

\subsection{Loss Function}
In most LRRG models, the cross-entropy loss is widely adopted to optimize the quality of generated reports. It leverages the negative log-likelihood to maximize the probability of generating each report word token $r_t$ conditioned on the input prompt $X$, which encompasses contextual information and previously generated word tokens. The loss is formulated as:
\begin{equation}
\mathcal{L}_{G} = 
- \sum_{t=1}^{L} 
\log \left(
p_{\theta}
\left(
r_t \mid 
 X
\right)
\right),
\end{equation}
where $\theta$ denotes the trainable parameters.

Building upon this formulation, several models~\citep{liu2025enhanced,bannur2023learning,wang2024hergen,liu2025hc,hou2023recap} incorporate an additional regularization loss, denoted as $\mathcal{L}_{reg}$, to impose additional constraints beyond the primary task objective, thereby enhancing the model performance.
\begin{equation}
\mathcal{L}_{Total}=\mathcal{L}_{G}+\mathcal{L}_{reg}.
\end{equation}
In particular, two representative forms of the above regularization loss have been widely adopted in LRRG models. One line of work~\citep{liu2025enhanced,bannur2023learning,wang2024hergen,liu2025hc} uses the contrastive learning loss $\mathcal{L}_{con}$ to enforce the cross-time and cross-modality feature alignment. Another line of research~\citep{liu2025hc,hou2023recap} introduces a temporal classification loss $\mathcal{L}_{class}$, which regularizes the model to capture accurate temporal progression patterns, thereby reinforcing temporal sensitivity during report generation.
Further details regarding these two loss functions are provided in Section~\ref{sec:Core Challenges and Approaches in Longitudinal Report Generation}.

In summary, LRRG methods are characterized by their reliance on powerful vision encoders and flexible text decoders, along with the specially designed loss function, collectively ensuring clinically reliable report generation.

\begin{table*}[t]
\centering
\scriptsize
\caption{Summary of representative LRRG models from 2023--2025, categorized by dataset composition and model architecture details.}
\label{tab:LRRG_summary}
\setlength{\tabcolsep}{4pt}%
\renewcommand{\arraystretch}{1.12}%
\resizebox{\textwidth}{!}{%
\begin{tabular}{l c c c c c c c c c c}
\toprule
\textbf{Model} & \textbf{Year} & \makecell{\textbf{Multi-}\\\textbf{View}} & \makecell{\textbf{Long-}\\\textbf{Only}} & \makecell{\textbf{Prior data}\\\textbf{Modalities}} & \makecell{\textbf{Multi-}\\\textbf{Prior Visits}} & \makecell{\textbf{Vision Encoder}} & \makecell{\textbf{Text Decoder}} & \makecell{\textbf{Contrastive}\\\textbf{Learning}} & \makecell{\textbf{Temporal}\\\textbf{Feature Fusion}} & \makecell{\textbf{Auxiliary}\\\textbf{Enhancement}\\\textbf{Module}} \\
\midrule
RECAP~\citep{hou2023recap} & 2023 & \checkmark &  & \textit{Rep/Img} &  & ViT & Transformer decoder &  & \checkmark & \checkmark \\
ULCX~\citep{zhu2023utilizing} & 2023 & \checkmark & \checkmark & \textit{Rep/Img} &  & ResNet101+Transformer & Transformer decoder &  & \checkmark &  \\
CCXRG~\citep{serra2023controllable} & 2023 &  &  & \textit{Img} &  & Faster R-CNN & Transformer decoder &  & \checkmark &  \\
BioVil-T~\citep{bannur2023learning} & 2023 &  &  & \textit{Rep/Img} &  & ResNet+Transformer & CXR-BERT & \checkmark & \checkmark &  \\
LDSSR~\citep{nicolson2024longitudinal} & 2024 & \checkmark &  & \textit{Rep} &  & CvT & Transformer decoder &  & \checkmark &  \\
HERGen~\citep{wang2024hergen} & 2024 &  &  & \textit{Img} & \checkmark & CvT & GPT2 & \checkmark & \checkmark & \checkmark \\
Libra~\citep{zhang2024libra} & 2024 &  &  & \textit{Img} &  & RAD-DINO & Meditron &  & \checkmark &  \\
EVOKE~\citep{miao2024evoke} & 2024 & \checkmark &  & \textit{Rep/Img} & \checkmark & ResNet101 & Transformer decoder & \checkmark &  &  \\
MAIRA-2~\citep{bannur2024maira} & 2024 & \checkmark &  & \textit{Rep/Img} &  & RAD-DINO & Vicuna &  & \checkmark &  \\
ERRG~\citep{santiesteban2024enhancing} & 2024 &  &  & \textit{Img} &  & ViT & MedLLaMA &  & \checkmark &  \\
MedVersa~\citep{zhou2024medversa} & 2024 & \checkmark &  & \textit{Img} &  & Swin Transformer & Llama-2-Chat &  & \checkmark &  \\
TiBiX~\citep{sanjeev2024tibix} & 2024 &  &  & \textit{Img} &  & VQ-GAN & CXR-BERT &  & \checkmark & \checkmark \\
M4CXR~\citep{park2024m4cxr} & 2024 & \checkmark &  & \textit{Rep/Img} &  & RAD-DINO & Mistral &  & \checkmark &  \\
HC-LLM~\citep{liu2025hc} & 2025 & \checkmark & \checkmark & \textit{Rep/Img} &  & Swin Transformer & MiniGPT4 & \checkmark & \checkmark & \checkmark \\
LLM-RG4~\citep{wang2025llm} & 2025 & \checkmark &  & \textit{Rep} &  & RAD-DINO & Vicuna &  & \checkmark &  \\
DDaTR~\citep{song2025ddatr} & 2025 & \checkmark &  & \textit{Rep/Img} &  & Swin Transformer & Transformer decoder &  & \checkmark &  \\
MLRG~\citep{liu2025enhanced} & 2025 & \checkmark &  & \textit{Rep/Img} & \checkmark & RAD-DINO & DistilGPT2 & \checkmark & \checkmark &  \\
RADAR~\citep{hou2025radar} & 2025 &  &  & \textit{Rep/Img} &  & SigLIP & Phi3-mini &  & \checkmark &  \\
STREAM~\citep{yang2025spatio} & 2025 & \checkmark &  & \textit{Rep/Img} & \checkmark & Swin Transformer & TinyLlama &  & \checkmark & \checkmark \\
PriorRG~\citep{liu2025priorrg} & 2025 & \checkmark &  & \textit{Img} &  & RAD-DINO & DistilGPT2 & \checkmark & \checkmark &  \\
Diff-RRG~\citep{yun2025diff} & 2025 & \checkmark & \checkmark & \textit{Rep/Img} &  & BiomedCLIP & BioMistral &  & \checkmark & \checkmark \\
\bottomrule
\end{tabular}
}
\end{table*}

\section{Core Challenges and Approaches in the LRRG task}
\label{sec:Core Challenges and Approaches in Longitudinal Report Generation}
Effective utilization of longitudinal data in the RRG task requires addressing several key challenges, which can be broadly categorized into three aspects:

\begin{itemize}
    \item \textbf{Missing or Inconsistent Longitudinal Data:}~Longitudinal records are often incomplete or irregular, requiring models to handle missing entries and uneven temporal distributions.
    \item \textbf{Feature Space Misalignment Across Time Steps:}~Representations from different time points and modalities can vary considerably, making effective alignment crucial for coherent temporal fusion.
    \item \textbf{Temporal Feature Integration:}~Strategies are required to fuse prior and current features for robust temporal reasoning and consistency.
\end{itemize}

Building on these challenges, this section reviews existing methods, summarizes strategies for addressing them, and further discusses auxiliary enhancement techniques designed to improve the exploitation of longitudinal data, as illustrated in Table~\ref{tab:LRRG_summary}.

\subsection{Handling Missing or Inconsistent Longitudinal Data}

Incorporating longitudinal data requires processing input from multiple time points. However, the availability of prior studies varies significantly across patients, and in some cases, historical data—defined here as either prior Chest X-ray images or prior radiology reports—may be entirely absent.

To mitigate these issues and ensure uniform input dimensions, recent models have adopted one of the following four strategies.

\subsubsection{Zero Padding for Missing Prior Studies}Several models~\citep{serra2023controllable,wang2025llm,sanjeev2024tibix,song2025ddatr} explicitly substitute zero vectors in cases where no historical data are available. This approach addresses missing longitudinal information while aligning input sequences to a fixed length. In longitudinal settings, missing data may occur both in the sequence of prior studies (e.g., absent images or reports at certain time points) and in the associated clinical context (e.g., missing Indication or Comparison sections). As a result, zero vectors may appear not only at the end of a sequence but also between valid inputs, which can mislead autoregressive models that attend to all preceding tokens.

To mitigate this issue, \textit{TiBiX}~\citep{sanjeev2024tibix} arranges valid inputs at the beginning of the sequence and consistently places zero-padded entries at the end. This layout minimizes the adverse effects of zero-padding and facilitates the construction of attention masks. Nevertheless, in models that explicitly fuse current and prior features, zero vectors may still introduce representational noise and degrade performance.

To address this, \textit{DDaTR}~\citep{song2025ddatr} modifies the gating mechanism by introducing a binary gating parameter $\alpha$, set to 1 when a prior examination is available and to 0 otherwise. This mechanism ensures that the final combined feature representation remains unaffected by missing inputs.

\subsubsection{Special Token-Based Representation for Absent Inputs}Unlike zero vectors, which do not explicitly signal the absence of information, some models~\citep{liu2025enhanced,bannur2023learning} introduce learnable placeholder tokens at the locations corresponding to missing data. These tokens serve not only as structural fillers but also as semantic indicators of missing information, thereby enabling the model to recognize and adapt to data absence more effectively.

\textit{BioVil-T}~\citep{bannur2023learning} bypasses the cross-temporal fusion module when the prior image is unavailable. It inserts a learnable placeholder token $\mathbf{P}^{\text{miss}} \in \mathbb{R}^{D_{\mathrm{img}}}$ into the current data feature, thereby preserving the input structure while explicitly signaling the absence of longitudinal information.

However, using a single generic placeholder token for all types of missing inputs may be insufficient for capturing the semantic distinctions among different absent components. Many models incorporate different sections of the radiology report (e.g., Indication, Prior Findings) into the input prompt. To address this, \textit{MLRG}~\citep{liu2025enhanced} introduces distinct placeholder tokens for different missing components, [NHI] for missing Indication and [NHPR] for missing Prior Findings, thus allowing the model to explicitly recognize which part of the clinical context is absent.

  \begin{figure*}[t]
    \centering
    \includegraphics[width=\linewidth]{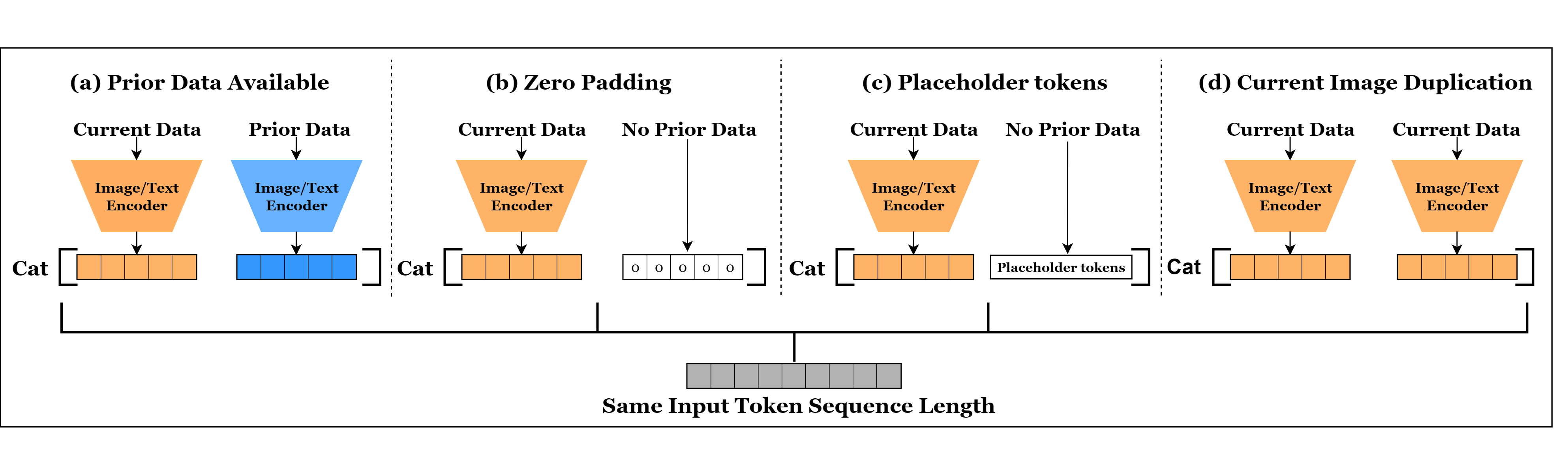}
    \caption{Strategies for unifying input sequence length in the LRRG task: (a) Using actual prior data when available; (b) Applying zero-padding when prior data is missing; (c) Placeholder tokens to indicate data absence; (d) Duplicating current data as a proxy for the missing prior input. All strategies ensure the same input token sequence length, enabling consistent model input formatting.}
    \label{fig:Strategies to deal with missing data}
\end{figure*}

\subsubsection{Prior Image Imputation via Current Image Duplication}While placeholder tokens can inform the model of missing data, their learned feature representations may not accurately reflect the distribution of real inputs. To address this discrepancy, several models~\citep{zhang2024libra,hou2025radar} replicate the current image as a proxy for the missing prior image, thereby maintaining consistency in feature distribution. However, this strategy risks misleading the model into interpreting identical images as distinct observations from different time steps, introducing temporal ambiguity.

To mitigate this issue, \textit{Libra}~\citep{zhang2024libra} introduces a learnable bias term that helps the model distinguish the true prior image from the dummy one based on its similarity to the current image. When the two images are highly similar, as in cases where a dummy prior is used, the bias is amplified, while in cases of low similarity, the bias is downweighted. This design enables the model to differentiate true temporal progression from duplicated or artificial priors, thereby mitigating hallucinations and improving clinical reliability.

\subsubsection{Using Pseudo Prior Reports for Prompting}While the first three strategies in Fig.~\ref{fig:Strategies to deal with missing data} operate at the input representation level to structurally compensate for absent data, the final approach instead addresses missing priors at the training strategy level by reusing previously generated reports as pseudo historical inputs. \textit{LDSSR}~\citep{nicolson2024longitudinal} proposes constructing continuous mini-batches of temporally ordered studies from the same patient, where the report generated in the preceding mini-batch is incorporated as a pseudo prior prompt for the current study.

This strategy effectively simulates real-world scenarios in which historical reports are unavailable, while encouraging the model to condition on its own prior outputs. As a result, the model learns to exploit longitudinal context even in the absence of authentic prior documentation.

Taken together, handling missing or inconsistent longitudinal data remains an open challenge, as substitutes for absent inputs often fail to convey clinically meaningful information. Emerging approaches that incorporate pseudo prior reports offer a more realistic means of modeling temporal continuity and point toward a promising direction for future advances in the LRRG task.
\subsection{Cross-Time and Cross-Modal Feature Alignment}

Unlike conventional single-image report generation, which relies on a single time point and fixed modalities, longitudinal modeling introduces variability across temporal and modality dimensions. This heterogeneity, stemming from variations in imaging protocols, modality combinations, and temporal intervals, complicates cross-time and cross-view integration. To address these challenges, most approaches employ contrastive learning. The following sections review contrastive learning strategies and their applications in the LRRG task.

\subsubsection{Contrastive Learning Foundations}Contrastive learning~\citep{radford2021learning} seeks to map semantically similar inputs closer in the embedding space while pushing dissimilar ones apart. For two modalities (or views) with $A$ samples each, $A^2$ feature pairs can be formed, but only $A$ are true positives. The objective maximizes similarity among positives while minimizing it for the remaining $A^2 - A$ negatives. This principle has proven highly effective in vision–language generation (VLG), where robust cross-modal alignment is essential.
A widely adopted implementation of this principle is the InfoNCE loss~\citep{oord2018representation}, which can be formulated as:
\begin{equation}
\mathcal{L}_{com} = -\frac{1}{B} \sum_{j=1}^{B} 
\log \frac{\exp \left( \text{sim}(q_j, k_j^+) / \tau \right)}
{\sum_{i=1}^{B} \exp \left( \text{sim}(q_j, k_i) / \tau \right)},
\end{equation}
where B is the batch size, the $q_j$ represents the anchor sample, $k_j^+$ is its corresponding positive match, and $k_i$ refers to all candidate samples within the batch. The similarity function $\text{sim}(\cdot)$ (typically cosine similarity) measures semantic consistency, $\tau$ is a temperature parameter.

\begin{figure*}[!t]
    \centering
    \includegraphics[width=\linewidth]{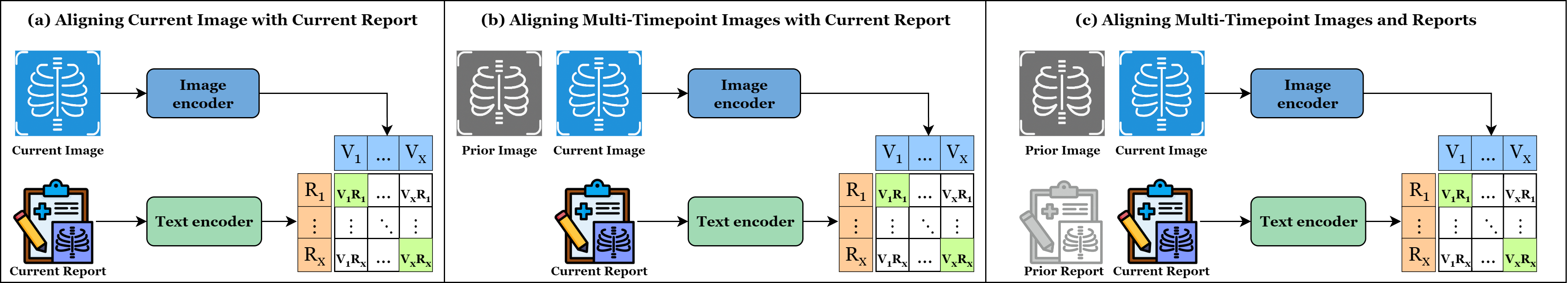}
    \caption{Illustration of different alignment strategies in radiology report generation. (a) Aligning the current image with the current report. (b) Aligning multi-timepoint images with the current report. (c) Jointly aligning multi-timepoint images and reports.}
    \label{fig:Feature alignment}
\end{figure*}

As a specialized subset of VLG, traditional RRG methods~\citep{sha2025contrastive} treat the visual and textual features corresponding to the same case within a batch as positive pairs, as shown in Fig.~\ref{fig:Feature alignment}(a). This concept can be naturally extended to the LRRG task, where contrastive learning aligns semantically consistent features between the global report representation and its corresponding fused cross-time feature visual representation within each batch. In this setting, temporal visual features and textual descriptions serve as cues for identifying positive pairs, as shown in Fig.~\ref{fig:Feature alignment}(b).

\subsubsection{Batch-Level Alignment across Studies}As discussed above, batch-level contrastive learning seeks to align global visual and textual representations within each training batch. Several recent studies~\citep{bannur2023learning, liu2025enhanced} have further refined this paradigm.

While global contrastive objectives facilitate coarse image–report alignment, they often overlook region–phrase correspondences that are critical for capturing clinically significant details. To mitigate this limitation, \textit{BioVil-T}~\citep{bannur2023learning} introduces additional local contrastive objectives, which compute token-level pairwise cosine similarities between visual and textual representations rather than relying solely on global features. These token-level similarities are optimized jointly using InfoNCE loss, encouraging fine-grained alignment.

Most existing batch-level contrastive learning methods 
specifically treat each image-report pair from the same case as a single positive sample. 
However, this design overlooks the fact that report contents may remain consistent across multiple visits within the same batch, leading to the false negative sample problem. 
To mitigate this limitation, \textit{MLRG}~\citep{liu2025enhanced} introduces a supervision signal \( \mathbf{p}^{g}_{i,j} \), defined as:
\begin{equation}
\mathbf{p}^{g}_{i,j} = 
\frac{\mathds{I}_{\text{identical}}\left(r^{\text{cur}}_i, r^{\text{cur}}_j\right)}
{\sum_{k=1}^{B} \mathds{I}_{\text{identical}}\left(r^{\text{cur}}_i, r^{\text{cur}}_k\right)}.
\end{equation}
Here, \( \mathds{I}_{\text{identical}}(\cdot,\cdot) \) denotes an indicator function 
that assigns positive labels not only to the report from the same visit but also 
to other reports that share identical content. Consequently, the positive samples of the vision representation will not be restricted to a single corresponding report but to all semantically identical reports within the batch. 
This supervision signal is then employed to construct a bidirectional InfoNCE-based 
contrastive objective, which jointly optimizes the predicted image-to-text and 
text-to-image similarity distributions \( q^{v2r} \) and \( q^{r2v} \). The image-to-text similarity \( q^{v2r} \) is computed as:
\begin{equation}
\mathbf{q}^{v2r} = \frac{\exp\left( \bar{\mathbf{V}} \cdot \bar{\mathbf{R}}^\top / \tau_2 \right)}{\sum_{j=1}^{B} \exp\left( \bar{\mathbf{V}} \cdot \bar{\mathbf{R}}_j^\top / \tau_2 \right)},
\end{equation}
where \( \bar{\mathbf{V}} \) and \( \bar{\mathbf{R}} \) denote global visual and textual representations. The text-to-image probability \( \mathbf{q}^{r2v} \) is defined analogously.
The final contrastive alignment loss is defined as:
\begin{equation}
\mathcal{L}_{con}^{MLRG} = -\frac{1}{B} \sum_{i=1}^{B} \left( p_i^g \log q_i^{v2r} + p_i^g \log q_i^{r2v} \right).
\end{equation}

\subsubsection{Patient-Level Alignment across Time}In contrast to the aforementioned methods that perform contrastive learning across data from different studies within the same batch, it is worth noting that a single batch may contain samples from multiple patients. This inter-patient variability often dominates over the subtle temporal differences that exist within the same patient, thereby limiting the model’s ability to capture true longitudinal variations. To address this issue, another line of work~\citep{wang2024hergen,liu2025hc} introduces contrastive objectives across multiple time points and corresponding reports of the same patient, as illustrated in part (c) of Fig.~\ref{fig:Feature alignment}. This intra-patient contrastive paradigm encourages the model to more effectively capture temporal discrepancies and disease progression patterns in longitudinal data.

\textit{HERGen}~\citep{wang2024hergen} introduces a global contrastive loss to align the visual and textual representations across different time points within the same patient. Specifically, the model applies mean pooling over image embeddings to construct a vision set $\left\{ v_s \right\}_{s=1,2,\ldots, N_B}$ and encodes the corresponding ground-truth reports to form a text set $\left\{ r_s \right\}_{s=1,2,\ldots, N_B}$, where $N_B$ denotes the number of studies. The same time point image-report pair will be treated as the positive sample, while the remaining pairs serve as negatives. The similarity distributions between matched visual and textual representations within these two sets are then optimized via an InfoNCE-based loss. This kind of contrastive objective mitigates inter-patient variability. However, intra-patient feature variance remains large enough to obscure features related to temporal progression.

In contrast, \textit{HC-LLM}~\citep{liu2025hc} further decomposes the current and prior visual $(v_c, v_p)$ and textual $(r_c, r_p)$ features into two categories: time-shared $(v^{sh}_c, r^{sh}_c, v^{sh}_p, r^{sh}_p)$ and time-specific $(v^{sp}_c, r^{sp}_c, v^{sp}_p, r^{sp}_p)$, in order to obtain more fine-grained and temporally sensitive feature representations. The detailed classification procedure is described in the Auxiliary Enhancement Modules section. As time-shared features are designed to remain invariant across different time points, the average time-shared features can be expressed as:
\begin{equation}
v^{sh} = \frac{v^{sh}_c + v^{sh}_p}{2}, \quad r^{sh} = \frac{r^{sh}_c + r^{sh}_p}{2}.
\end{equation}
These are concatenated with the time-specific features to form the final image and text sequences:
\begin{align}
\tilde{v} = [v^{sh}, v_p^{sp}, v_c^{sp}], \quad \tilde{r} = [r^{sh}, r_p^{sp}, r_c^{sp}].
\end{align}
In this setup, modality pairs sharing the same temporal progression characteristics (e.g., prior-specific vision and texture features) are treated as positive samples, while the remaining pairs serve as negatives. Specifically, the InfoNCE loss is adopted to guide the similarity distribution between positive and negative pairs.

In summary, contrastive learning has been widely employed to align features across time and modality in the LRRG task. Although existing methods vary in defining positive pairs and structuring alignment losses, most of them still lack fine-grained feature alignment specifically designed to model temporal progression, which represents a potential direction for future improvement.

\subsection{Longitudinal Feature Fusion Strategies}
While temporal and cross-modal alignment is essential, the effectiveness of longitudinal radiology report generation ultimately hinges on the fusion of the aligned information. Radiology reports contextualize current findings by selectively weighting prior observations, requiring models to determine both the historical cues to integrate and the degree to which they should be emphasized.

In the following, we review fusion strategies employed in recent models to integrate current and prior inputs after alignment. According to their underlying mechanisms, we classify these strategies into three categories: (i) concatenation-based methods; (ii) gated mechanism-based methods; and (iii) attention-based fusion methods.

\begin{figure}[!t]
    \centering
    \includegraphics[width=\columnwidth]{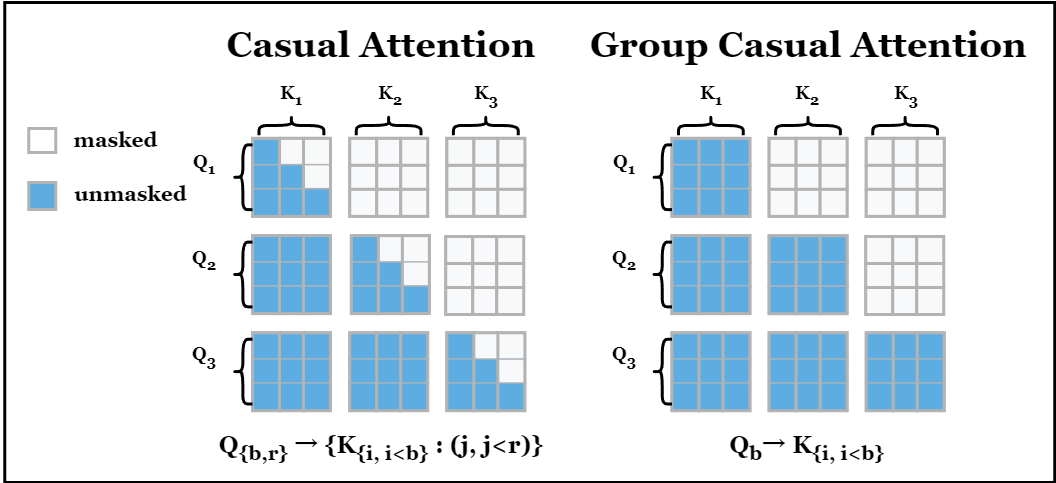}
    \caption{Comparison between causal attention and Group causal attention~\citep{wang2024hergen}, highlighting the number of tokens each query can attend to. Group causal attention enables interaction with both preceding tokens and all tokens within the same group, resulting in a broader attention scope.}
    \label{fig:casual-attention difference}
\end{figure}

\subsubsection{Concatenation-Based Methods}Without designing a specific feature fusion mechanism, several methods employ direct concatenation of features for temporal fusion. Despite their simplicity, such strategies remain popular due to their ease of implementation and adaptability. Among these, two methods~\citep{serra2023controllable,wang2024hergen} stand out for their distinct design goals and fusion scopes.

\textit{CCXRG}~\citep{serra2023controllable} performs region-level concatenation of features across time. A Faster R-CNN detects 36 anatomical regions from both current and prior images, yielding two sets of region-specific features:
\begin{align}
V_{\mathrm{current}} = \{\vec v_{n}^{curr}\}_{n=1}^N, \quad
V_{\mathrm{prior}} = \{\vec v_{n}^{prior}\}_{n=1}^N,
\end{align}
where \( \vec v_{n}^{curr} \) and \( \vec v_{n}^{prior} \) denote the features of the \( n \)-th region at current and prior time points, N is the total number of regions.
To enable aligned fusion, corresponding regions are concatenated and passed through a longitudinal projection module \( f \), composed of two fully connected layers and batch normalization, producing joint region-level representations.
This design also supports region-targeted generation. For a specified subset of regions \( A_{\mathrm{target}} \subseteq \{a_1, \dots, a_N\} \), only the corresponding region pairs are processed, while others are zero-padded:
\begin{align}
\vec v_{n}^\mathrm{joint} =
\begin{cases}
f\left([\vec v_{n}^{curr},\, \vec v_{n}^{prior}]\right), & a_n \in A_{\mathrm{target}}, \\[4pt]
f\left([\vec 0,\, \vec 0]\right).             & \text{otherwise}.
\end{cases}
\end{align}

\textit{HERGen}~\citep{wang2024hergen}, in contrast, extends concatenation across a longer temporal window. It first applies mean pooling to image features at each of five time points, yielding global representations per visit. These are then concatenated to form a temporally aggregated visual embedding.

\subsubsection{Gated Mechanism-Based Methods}Different from simple concatenation-based fusion, recent approaches~\citep{hou2023recap,song2025ddatr} have evolved toward gated mechanisms that adaptively regulate cross-temporal and cross-modal feature integration, thereby enhancing the focus on clinically relevant information while suppressing redundancy.

\textit{DDaTR}~\citep{song2025ddatr} introduces a Dynamic Difference-aware Module (DDAM) to reduce discrepancies between current and prior image features. 
Specifically, the Dynamic Difference-aware Module (DDAM) first computes the pixel-wise difference between the current and prior features, which is then average-pooled and passed through a sigmoid to generate a difference-aware weight map. This weight adaptively re-scales both current and prior features via element-wise multiplication, thereby highlighting regions of change. The resulting difference-enhanced features act as a dynamic gating factor: when the discrepancy between current and prior images is large, the gating mechanism emphasizes the difference features more strongly in combination with the current representation.

\textit{RECAP}~\citep{hou2023recap} generalizes this idea by introducing dual gating mechanisms. The first gate adaptively fuses the observation-aware hidden state from the current radiology and observations with the progression-aware hidden state from prior records, where the gating factor is derived from the current-image representation. The second gate balances two prediction streams: one from the attention-based decoder and the other from the disease-progression graph. The graph encodes structured links between observations and fine-grained spatial or temporal attributes, while the gating weight, computed from the fused hidden state, determines the extent to which these graph-derived attributes shape the final output.

\subsubsection{Attention-Based Fusion Methods}Attention mechanisms, particularly within Transformer architectures~\citep{vaswani2017attention}, have become a cornerstone for multimodal feature integration due to their ability to capture long-range dependencies and flexible contextual relationships.

In vision-language generation (VLG), models such as \textit{Flamingo}~\citep{alayrac2022flamingo} tokenize both visual and textual inputs into a unified sequence, enabling self-attention layers to learn global cross-modal interactions and thereby enhance generation quality. This paradigm has influenced the LRRG task, where the key challenge shifts to effectively incorporating temporally distributed historical information.

Recent LRRG models extend attention-based fusion to handle both current and prior inputs. These methods differ primarily in how they formulate query vectors during attention computation, which determines the focus of contextual integration. We group them into two main approaches:

\noindent\textbf{Token-Level Self-Attention Fusion.}~Self-attention has been widely adopted in the LRRG methods~\citep{nicolson2024longitudinal,bannur2024maira,bannur2023learning,sanjeev2024tibix,park2024m4cxr,wang2024hergen}. In this setting, each token serves as a query attending to all others, enabling temporal fusion across time points.

Most approaches input visual tokens directly into a large language model and rely on standard self-attention to implicitly capture longitudinal dependencies. While some methods introduce architectural modifications to better reflect the structured nature of sequential radiology data and address temporal integration more explicitly.

\textit{BioVil-T}~\citep{bannur2023learning} employs a token-level fusion mechanism, where current and prior image features are concatenated with positional and temporal embeddings before being propagated through stacked self-attention layers. Within this joint sequence, self-attention updates the current-image tokens using information from both themselves and the prior-image tokens, thereby embedding longitudinal context directly into the representation of the current examination. After the final layer, only the updated current image tokens are retained and concatenated with the original current image embedding to form the final visual representation. This design effectively fuses prior features while simultaneously reducing redundancy and maintaining clinical focus on the present examination.

\textit{HERGen}~\citep{wang2024hergen} represents another line of work that modifies the attention mechanism by proposing Group Causal Attention. Unlike standard causal attention, where each vision patch token can only attend to earlier patches and therefore lacks full spatial context, Group Causal Attention enables each token to attend freely to all vision tokens within the same time-point group, regardless of their sequential order, shown in Fig.~\ref{fig:casual-attention difference}. This design preserves comprehensive spatial dependencies within each image, alleviates the fragmentation of visual evidence caused by sequential tokenization, and supports both intra-image and cross-time interactions while maintaining autoregressive constraints. As a result, the model is better equipped to capture longitudinal visual token features.

\begin{figure}[!t]
    \centering
    \includegraphics[width=\columnwidth]{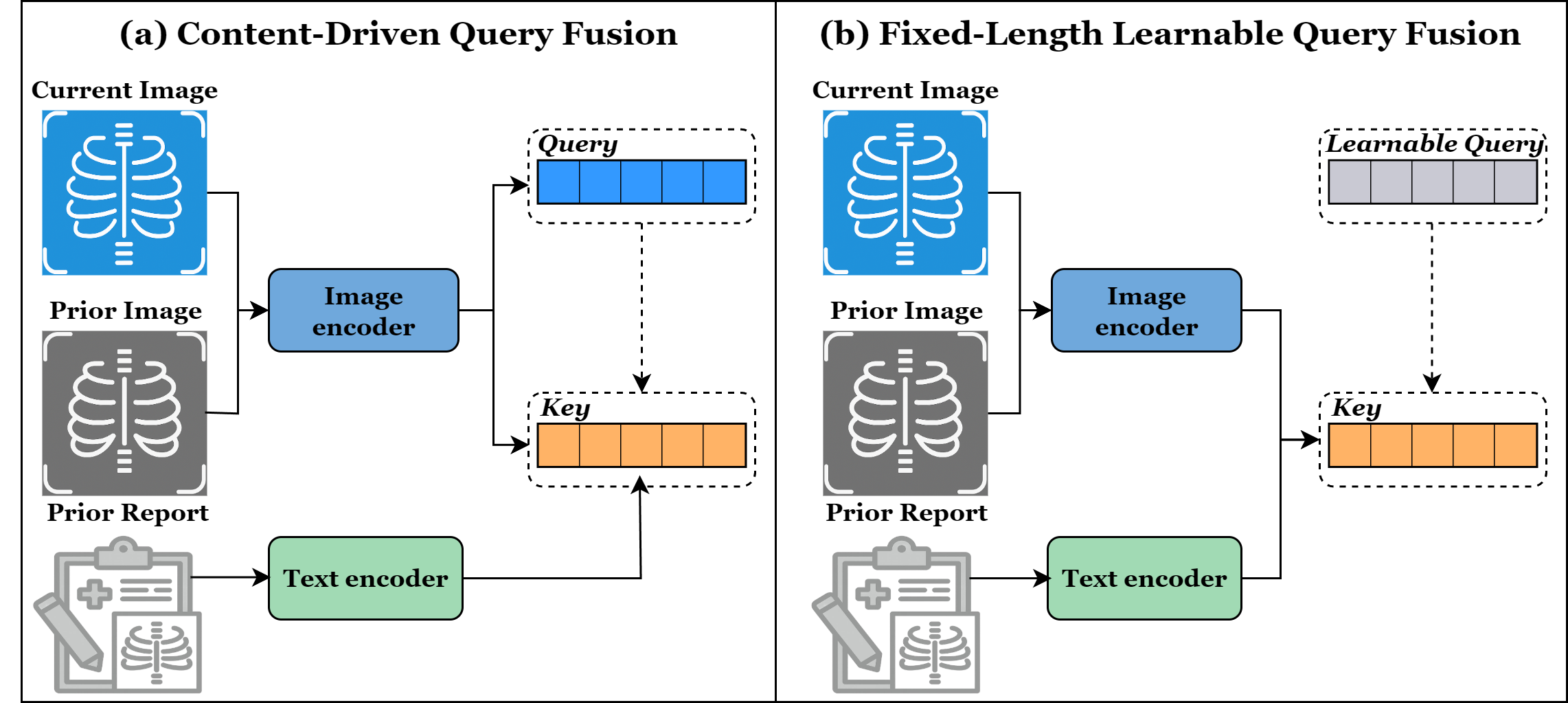}
    \caption{Comparison of two cross-attention paradigms for the LRRG task. (a) Content-Driven query fusion, where the current image feature serves as the query and interacts with prior image and report features as keys. (b) Fixed-length learnable query fusion, where a learnable query vector attends to current and prior data features encoded as keys.}
    \label{fig:cross-attention difference}
\end{figure}

\noindent\textbf{Query-Based Cross-Attention Fusion.}~While self-attention enables rich token-level interactions, its quadratic complexity becomes prohibitive for long input sequences—particularly in longitudinal scenarios where multiple time points and views must be processed jointly. To alleviate this burden, several models~\citep{zhu2023utilizing, liu2025enhanced, zhang2024libra, wang2025llm, santiesteban2024enhancing} introduce cross-attention mechanisms, which can be broadly categorized into two types, as illustrated in Fig.~\ref{fig:cross-attention difference}. Such designs facilitate temporal feature fusion while constraining the overall sequence length.

\noindent\textbf{(1) Fixed-Length Learnable Queries.}~One class of models introduces a set of fixed-length learnable query vectors~\citep{li2022blip} that perform cross-attention over variable-length input features. This produces a consistent number of output tokens regardless of input length.

For example, \textit{ERRG}~\citep{santiesteban2024enhancing} applies a separate Perceiver~\citep{jaegle2021perceiver} to each image. Each Perceiver processes the image independently, producing a fixed-length representation later consumed by a language model. However, this design neglects interactions between current and historical inputs during feature extraction.

In contrast, \textit{LLM-RG4}~\citep{wang2025llm} introduces a hierarchical fusion architecture using three Perceivers \( p_f \), \( p_l \), and \( p_t \), each attending to different modalities:
\begin{align}
\mathbf{h}_f &= \mathrm{Linear}(p_f(\mathbf{v}_f, \mathbf{V}')), \\
\mathbf{h}_l &= \mathrm{Linear}(p_l(\mathbf{v}_l, p_f(\mathbf{v}_f, \mathbf{V}'))), \\
\mathbf{h}_t &= \mathrm{Linear}(p_t(\mathbf{r}_t, p_f(\mathbf{v}_f, \mathbf{V}'))).
\end{align}
Here, \( \mathbf{V}' \) denotes fixed learnable queries, while \( \mathbf{v}_f \), \( \mathbf{v}_l \), and \( \mathbf{r}_t \) correspond to frontal-view, lateral-view, and prior report features, respectively. The output of \( p_f(\mathbf{v}_f, \mathbf{V}') \) serves as the query input for both \( p_l \) and \( p_t \), and the resulting representations \( \mathbf{h}_f \), \( \mathbf{h}_l \), and \( \mathbf{h}_t \) are subsequently concatenated to form the final multimodal embedding.Unlike methods that merely compress each input into an isolated representation, LLM-RG4 explicitly models inter-feature dependencies during fusion, thereby preserving complementary cues across views and textual priors.

\noindent\textbf{(2) Content-Driven Query Fusion.}~While fixed learnable queries yield compact global abstractions but sacrifice spatial correspondence, another line of work preserves spatial structure by employing current image tokens as queries to guide attention over auxiliary features, thereby enabling the selective integration of historical or multi-view information.

\textit{MLRG}~\citep{liu2025enhanced} introduces a multi-view longitudinal fusion (MLF) network, designating one current image view as the anchor and treating the remaining views and prior images as auxiliary. The anchor’s token sequence serves as the query, while other inputs are concatenated as keys and values.
This design enhances cross-view and cross-temporal integration; however, it inherently neglects the internal spatial dependencies within the anchor view, limiting its capacity to capture fine-grained intra-view relationships.

\textit{Libra}~\citep{zhang2024libra} addresses this by introducing the Temporal Fusion Module (TFM), which combines self-attention and cross-attention:
\begin{align}
T_{\text{curr}}^{\text{self}} &= \mathrm{LN}\left(v^{\text{curr}} + \mathrm{SelfAttn}(v^{\text{curr}})\right), \\
T_{\text{prior}}^{\text{self}} &= \mathrm{LN}\left(v^{\text{prior}} + \mathrm{SelfAttn}(v^{\text{prior}})\right), \\
T_{\text{img}}^{\text{cross}} &= \mathrm{LN}\left(T_{\text{curr}}^{\text{self}} + \mathrm{CrossAttn}(T_{\text{curr}}^{\text{self}}, T_{\text{prior}}^{\text{self}})\right).
\end{align}
This pipeline allows the model to preserve intra-image structure via self-attention and capture cross-temporal dependencies via guided cross-attention, thereby achieving a more balanced integration of local fidelity and longitudinal context.

Overall, cross-attention mechanisms are central to effective temporal and multi-view integration in the LRRG task. Fixed-length learnable queries offer computational efficiency, while content-driven queries retain spatial inductive bias and temporal specificity. Hierarchical fusion and modular attention blocks show promise in structuring clinically relevant dependencies, underscoring the importance of attention design in longitudinal contexts.

\subsection{Auxiliary Enhancement Modules}

While fusion strategies are central to longitudinal modeling, many models also integrate auxiliary modules to further improve temporal reasoning and contextual understanding. We categorize these enhancements into three groups.

\begin{figure}[!t]
    \centering
    \includegraphics[width=0.7\columnwidth]{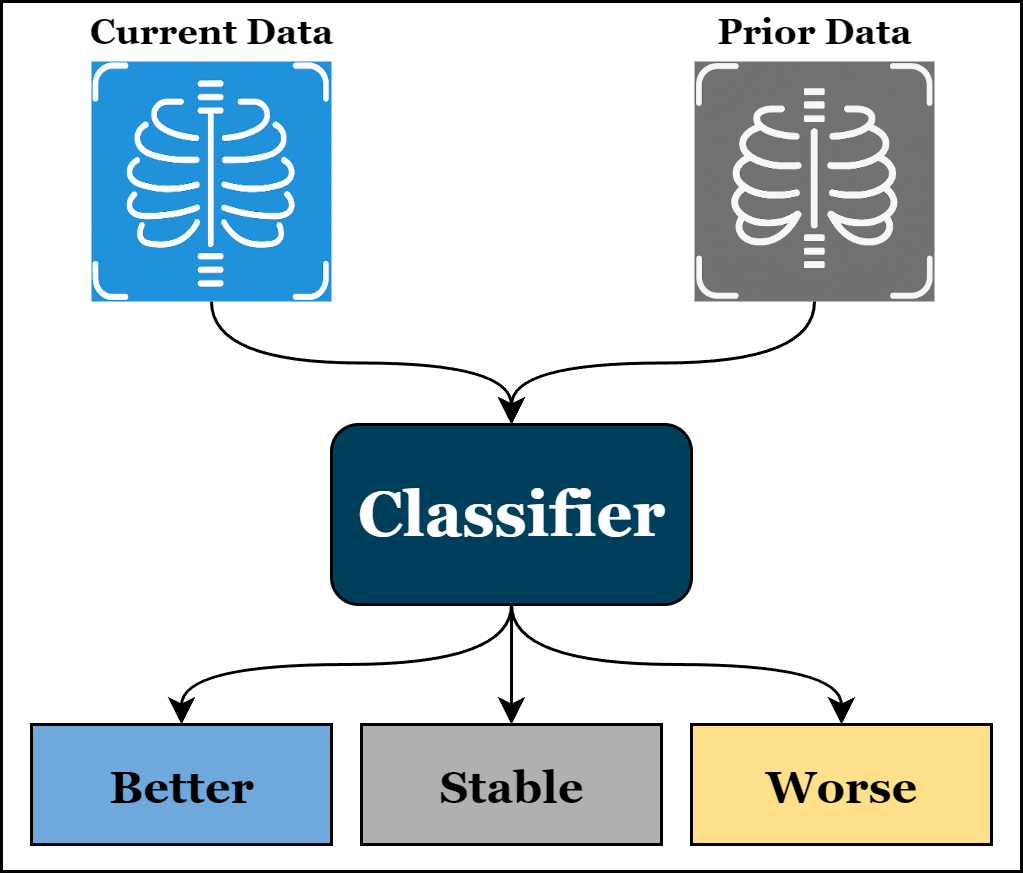}
    \caption{Temporal progression classification in RECAP~\citep{hou2023recap}. The model compares current and prior images and assigns one of three progression labels: better, stable, or worse.}
    \label{fig:temperal classification}
\end{figure}

\subsubsection{Temporal change classification}Effective modeling of longitudinal data requires mechanisms that enable explicit recognition of changes across time. While many approaches simply feed current and prior data into a shared architecture, this often makes it difficult for the model to learn the features that are most useful for capturing temporal changes. To address this, recent methods introduce classification-based strategies that make temporal dynamics more interpretable and structured.

\textit{HC-LLM}~\citep{liu2025hc} disentangles temporally invariant and variant components by classifying feature relationships into two categories: time-shared and time-specific. Given a current image \( I_c \) and a prior image \( I_p \), their visual features are extracted via a vision encoder \( f_{\mathrm{ve}} \), and the outputs at the \texttt{[CLS]} positions are projected into the text embedding space:
\begin{align}
v_c = W_v \cdot f_{\mathrm{ve}}(I_c)_{\texttt{[CLS]}}, \quad
v_p = W_v \cdot f_{\mathrm{ve}}(I_p)_{\texttt{[CLS]}},
\end{align}
where \( W_v \) is a learned projection matrix. Similarly, the generated current report \( \hat{R}_c \) and the historical report \( R_p \) are processed by a text encoder \( f_{\mathrm{tg}} \), using their \texttt{[CLS]} tokens as global representations:
\begin{align}
r_c = f_{\mathrm{tg}}(\hat{R}_c)_{\texttt{[CLS]}}, \quad
r_p = f_{\mathrm{tg}}(R_p)_{\texttt{[CLS]}}.
\end{align}
Each feature—visual or textual—is then passed through two branches: a shared encoder \( E_{sh} \) that captures time-invariant information, and a time-specific encoder (\( E_{sp}^c \) for current, \( E_{sp}^p \) for prior) that models temporal variations.
To enforce disentanglement, it uses the MSE loss to encourage shared features to remain consistent across time. While \textit{HC-LLM} distinguishes whether features are shared or specific across time, it does not quantify the direction of change.

\textit{RECAP}~\citep{hou2023recap} addresses this by leveraging progression labels from the Chest ImaGenome dataset~\citep{wu2021chest}, which annotate each ROI as \textit{better}, \textit{worse}, or \textit{stable}, shown in Fig.~\ref{fig:temperal classification}. RECAP reformulates this into an image-level classification task that predicts whether the current image \( I^c \) reflects a relative time progression change to the prior image \( I^p \).
The model extracts \texttt{[CLS]} tokens from both images and concatenates them to form a joint representation, which will be later used to predict the probability of the j-th progression $p(p_j)$:
\begin{align}
[\mathrm{CLS}] = [[\mathrm{CLS}]^p\;;\;[\mathrm{CLS}]^c], \qquad
p(p_j) = \sigma(W_j\, [\mathrm{CLS}] + b_j),
\end{align}
where \( W_j \) and \( b_j \) are learnable weight matrix and bias. The prediction is trained with a binary cross-entropy loss.

Together, these approaches enhance temporal modeling by explicitly classifying temporal relationships to provide stronger supervision for change detection.

\subsubsection{Retrieval‑Based Augmentation}Beyond classifying temporal changes, another line of work~\citep{hou2023recap,yang2025spatio} enhances longitudinal modeling by retrieving external knowledge to enrich internal representations. 

\textit{RECAP}~\citep{hou2023recap} constructs a relational graph based on entities and relationships extracted from auxiliary datasets~\citep{bannur2023learning,jain2021radgraph}. Temporal entities \( E^T \) (e.g., \textit{bigger}, \textit{decreased}) describe longitudinal trends, while spatial entities \( E^S \) (e.g., \textit{healed}, \textit{top}) encode local structural variations. The graph's node set is defined as 
\( N = \{ O, E^T, E^S \} \), where \( O \) denotes current observations derived using CheXbert. Edges are defined as \( D = \{ S, B, W, R_S, R_O \} \), with \( S \), \( B \), and \( W \) indicating the semantic relations \textit{Stable}, \textit{Better}, and \textit{Worse}. \( R_S \) links observations to spatial entities, while \( R_O \) connects current and prior observations.
To construct the graph, RECAP computes pointwise mutual information (PMI)~\citep{church1990word} between observation-relation pairs and candidate entities. For a given pair \( (o_i, r_j) \), where \( r_j \in D \), we select the top-\( K \) entities \( e_k^* \in E^T \cup E^S \) that achieve the highest PMI scores:
\begin{align}
\mathrm{PMI}(\bar{x}, \hat{x}) = \log \frac{p(\hat{x} \mid \bar{x})}{p(\hat{x})}, \quad
 \bar{x} = (o_i, r_j), \; \hat{x} = e_k^*.
\end{align}
This process yields a graph that captures co-occurrence patterns between observed findings and semantic knowledge. Although effective, \textit{RECAP} performs retrieval and reasoning at the image level without explicitly modeling region-level dynamics grounded in anatomical structures.

To address this, \textit{STREAM}~\citep{yang2025spatio} builds a region-of-interest (ROI) library from Chest ImaGenome~\citep{wu2021chest} and MIMIC-CXR~\citep{johnson2019mimic}. Each entry stores: visual features \( v_i \), anatomical organ region label \( l_i \), textual description \( s_i \), abnormality indicator \( a_i \), and a temporal identifier list \( c_i \), which shows corresponding historical ROIs:
\begin{align}
y_i = \{ v_i, l_i, s_i, a_i, c_i \}.
\end{align}
During inference, 29 ROIs are extracted from the input image. For each ROI region, the top-\( k \) most similar regions are retrieved. If prior data exist for both the input and a retrieved candidate, their historical ROIs are further compared, and similarity-based refinement is applied. Final selection proceeds by majority voting on the abnormality indicator \( a_i \), filtering by label compatibility, and ranking by similarity. The highest similarity region’s description \( s_i \) is used as a prompt for report generation.

\subsubsection{Time Gap Encoding}While retrieval-based methods incorporate external knowledge to model longitudinal dependencies, another complementary direction~\citep{wang2024hergen,sanjeev2024tibix} focuses on utilizing time gaps between studies.

\textit{TiBiX}~\citep{sanjeev2024tibix} introduces a simple yet effective mechanism: it encodes the interval between two consecutive studies as a temporal token, which is added to the input sequence to inform the model of the time elapsed between samples.

\textit{HERGen}~\citep{wang2024hergen} introduces \textit{time vocabulary}, whose length corresponds to the maximum temporal gap in the input longitudinal data, to encode relative temporal positions. For each image taken at time \( T_j^{(i)} \), its offset from the reference time \( T_0^{(i)} \) is computed:
\begin{align}
T_j^{\prime(i)} = T_j^{(i)} - T_0^{(i)}.
\end{align}
These offsets are mapped to embedding vectors:
\begin{align}
p_j^{(i)} = \mathrm{Embedding}(T_j^{\prime(i)}) \in \mathbb{R}^{1 \times F'}.
\end{align}
Each embedding is added to the corresponding vision tokens, and the temporally enriched sequences are concatenated to form the final input.

In summary, auxiliary enhancement modules offer critical support for longitudinal modeling by enriching temporal representations beyond core fusion mechanisms. Through explicit change classification, structured knowledge retrieval, and effective encoding of time intervals, these methods improve the model’s capacity to reason about disease progression and temporal context.

\begin{figure}[!t]
    \centering
    \includegraphics[width=\columnwidth]{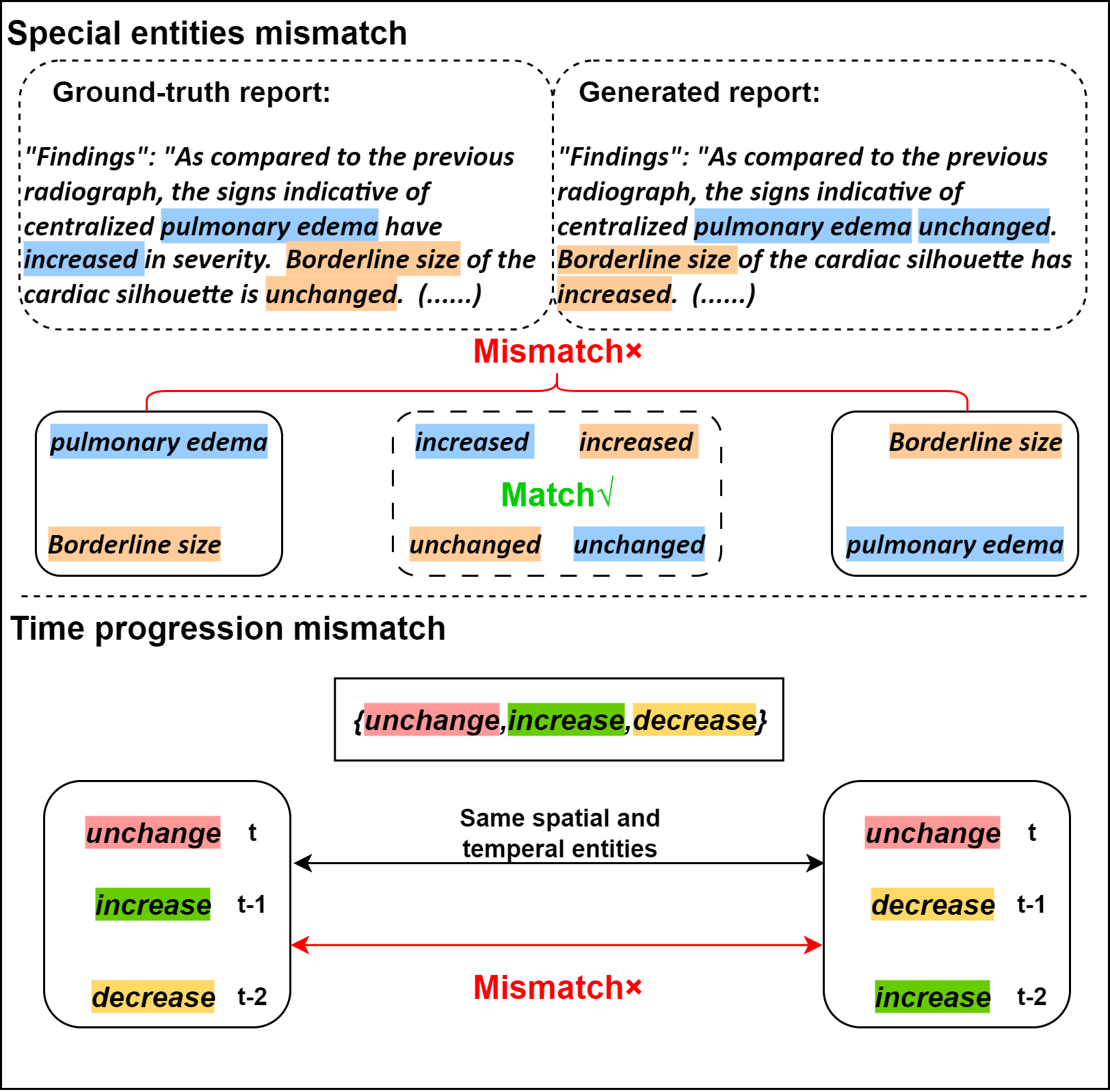}
    \caption{Examples of mismatch problems in radiology report evaluation. 
The upper part illustrates spatial entity mismatches, where temporal entities match but are attached to incorrect spatial entities. 
The lower part illustrates time progression mismatches, where the same spatial and temporal entities are involved, but the sequence of temporal changes is incorrectly ordered. 
These cases demonstrate the limitation of simply matching temporal entities between the generated and reference reports.
}
    \label{fig:mismatch problem}
\end{figure}

  \begin{figure*}[!t]
    \centering
    \includegraphics[width=\linewidth]{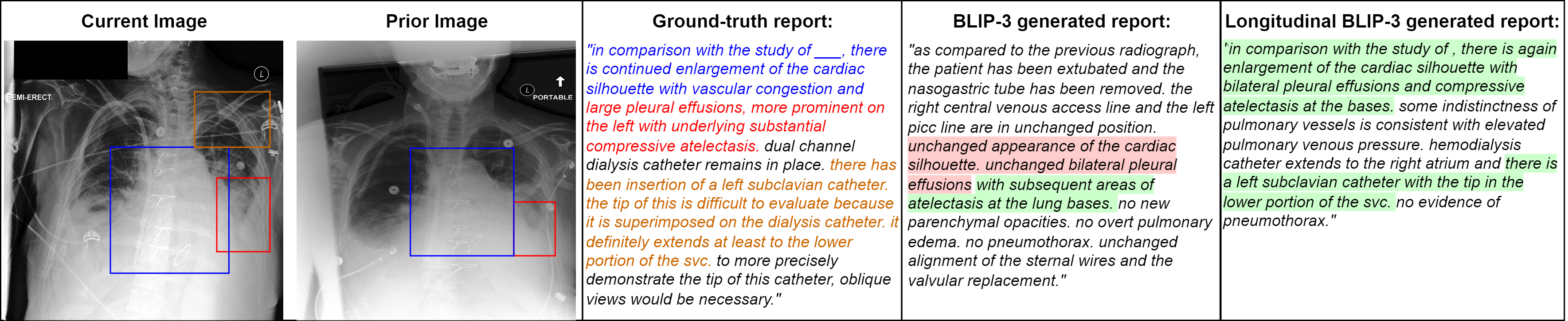}
    \caption{Case study on the MIMIC-CXR test set comparing BLIP-3 (single-image) with its longitudinal variant. “BLIP-3” denotes the single-image setting, while “Longitudinal BLIP-3” denotes the longitudinal setting. Annotations link each ground-truth report observation to corresponding image bounding boxes in the current and prior images, with matched pairs shown in the same color. In the generated report, correct findings are highlighted in green; hallucinated observations are marked in red.
}
    \label{fig:Case study}
\end{figure*}

\section{Evaluation Metrics}
\label{sec:Evaluation Metrics}

This section reviews the primary evaluation metrics used for the RRG task, which can be broadly categorized into two types. Lexical metrics serve as standard measures of surface-level similarity, reflecting fluency and linguistic overlap between generated and reference reports. Clinical metrics provide task-specific evaluation by assessing whether generated reports capture correct clinical findings and relations, thereby reflecting factual accuracy and diagnostic reliability. Both categories are presented in Tables~\ref{tb:NLG comparison MIMIC}, \ref{tb:CE comparison MIMIC}, \ref{tb:New CE metric MIMIC}, and~\ref{tb:Longitudinal dataset comparison} to illustrate the performance of LRRG methods.

In addition, given the strong performance of LLM-based metrics, which leverage large models’ semantic understanding and medical knowledge to move beyond n-gram overlap while simultaneously assessing linguistic quality, factual correctness, and clinical relevance, several examples of LLM-based metrics that incorporate longitudinal-related scores are also introduced. Furthermore, metrics specifically designed for longitudinal evaluation are included in this section, which aim to determine whether the generated reports correctly capture the temporal progression between examinations.

Based on the description above, the evaluation of RRG models can be broadly divided into four categories:

\subsection{Lexical Metrics}
To assess surface-level text similarity, lexical metrics are widely used in most RRG tasks. Bleu~\citep{papineni2002bleu} computes the geometric mean of modified 
n-gram precisions; METEOR~\citep{banerjee2005meteor} combines semantic precision and recall using an alignment-based harmonic score; ROUGE-L~\citep{lin2004rouge} measures the F-score based on the longest common subsequence.

\subsection{Clinical Metrics}
To minimize redundancy and emphasize clinically relevant terminology, clinical metrics have become essential for evaluating the performance of different methods. The key clinical metrics are as follows:

\subsubsection{CheXbert-based Metrics}CheXbert-based metrics~\citep{smit2020chexbert} classify 14 radiological observations into four categories (Positive, Negative, Uncertain, Blank), following the CheXpert~\citep{irvin2019chexpert} protocol. Model performance is evaluated using these metrics, including Recall, Precision, and F1 scores, which have both macro and micro forms. In addition, CheXbert vector similarity~\citep{smit2020chexbert, yu2023evaluating} is employed to quantify semantic alignment through cosine similarity between the generated and reference report embeddings.

\subsubsection{RadGraph-based Metrics}RadGraph-based metrics build on RadGraph~\citep{jain2021radgraph}, which annotates clinical entities and their relations. RadGraph-F1~\citep{jain2021radgraph} measures structural agreement by calculating the F1 score for entity and relation matches separately between generated and reference reports, followed by averaging the results. The  $RG_{ER}$ metric~\citep{delbrouck2022improving} matches the generated report and the ground-truth report not only based on the entities but also on the corresponding relationships.

\subsubsection{RadCliQ}RadCliQ~\citep{yu2023evaluating} employs a linear regression model to integrate multiple evaluation metrics, which better aligns with radiologists’ clinical assessments. The version 0 implementation of RadCliQ is commonly used to evaluate the performance of different report generation models.

Having outlined the metrics used to compare different methods, it is worth noting that these approaches generally overlook the longitudinal dimension. To address this gap, we further introduce metrics that are relevant for assessing longitudinal aspects.

\subsection{Large Language Model (LLM)-based Metrics}
LLM-based evaluation metrics align more closely with expert assessment while offering finer-grained, more interpretable feedback than conventional n-gram measures.

Given the strong performance of online large language models (LLMs) such as GPT-4o, a number of studies~\citep{huang2024fineradscore,jiang2025clear} have employed them for radiology report evaluation.
FineRadScore~\citep{huang2024fineradscore} extends the existing evaluation framework by introducing a more granular taxonomy of clinical errors. Specifically, it classifies erroneous references to longitudinal information under a distinct error category termed “Invalid comparison''. Each discrepancy between the generated and reference reports is annotated with clinical severity and error type.  
Despite evaluating the overall comparison against prior studies, CLEAR~\citep{jiang2025clear} leverages LLM to extract two longitudinally related structured attributes: first occurrence and change. These two subscores assess whether the model correctly identifies new clinical observations and captures temporal progression across serial studies, enabling a fine-grained measurement of longitudinal alignment between generated and ground-truth reports.

However, the use of online models is often impractical due to patient privacy concerns. Consequently, another line of research~\citep{liu2024mrscore,ostmeier2024green,jiang2025clear,li2025radreason} has explored the use of offline LLMs, such as LLaMA, which better protect patient privacy and are more cost-effective. 
MRScore~\citep{liu2024mrscore} leverages a seven-dimensional framework jointly evaluating semantic correctness and linguistic quality. Among them, the “Clinical history” dimension explicitly examines whether generated reports correctly reference prior examinations, making it particularly relevant for longitudinal assessment.  
Beyond scalar evaluation, GREEN~\citep{ostmeier2024green} provides an interpretable taxonomy of six error subtypes. Notably, the f-option indicates missing comparisons with prior studies, while the e-option flags hallucinated comparisons, thereby reflecting a model’s ability to capture longitudinal information. 

\begin{figure}[!t]
    \centering
    \includegraphics[width=\columnwidth]{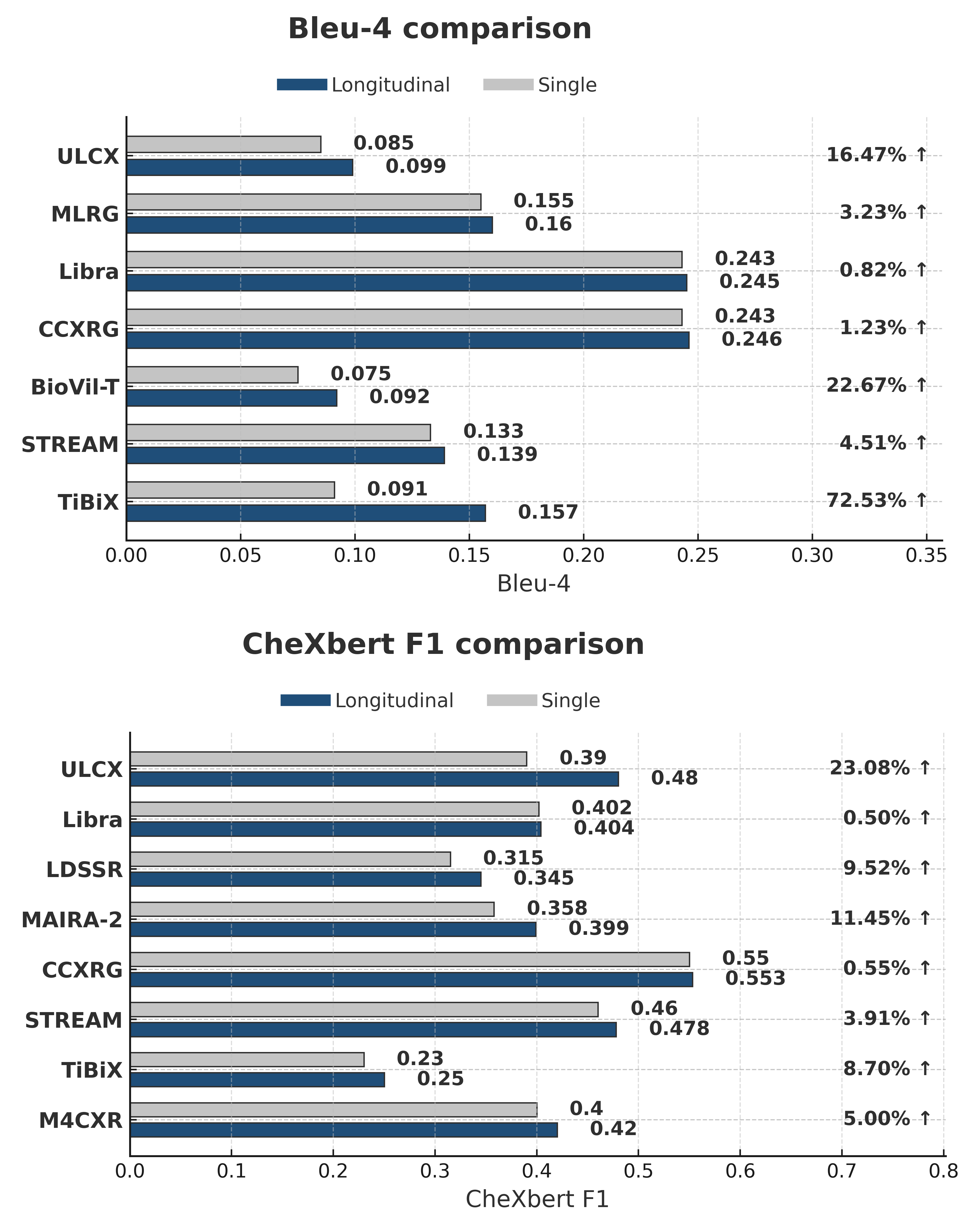}
    \caption{Visualization comparing different methods under the single-image vs. longitudinal settings, directly illustrating the performance gap between the two.}
    \label{fig:visual_single_vs_longitudinal}
\end{figure}

\subsection{Longitudinal-related Metrics}
All of the above LLM-based metrics include components for assessing longitudinal information, but their primary goal remains a comprehensive evaluation of overall model performance rather than dedicated measurement of temporal progression across time points. In the following subsection, we review metrics explicitly designed for longitudinal settings.

The Temporal F1~\citep{zhang2024libra} and TEM~\citep{bannur2023learning} are longitudinal-related metrics based on the temporal entity set provided by \textit{BioVil-T}~\citep{bannur2023learning}, which mainly contains different temporal changes to the extracted clinical entities. (e.g., bigger, change, cleared) . Both metrics calculate precision and recall based on the overlap of these temporal entities between the generated and ground-truth reports, in order to evaluate the model’s performance in capturing temporal progression information.

However, these methods have some drawbacks, shown in Fig.~\ref{fig:mismatch problem}. They extract temporal entities without ensuring spatial alignment, leaving some errors undetected. Moreover, correctness at a single time point does not guarantee accurate modeling of disease progression; metrics that evaluate multiple time points are needed to better reflect performance in capturing temporal dynamics.

To implement these functions, LUNGUAGESCORE
~\citep{moon2025lunguage} further extends evaluation to the sequential-report setting. It introduces the concept of TEMPORALGROUP, which not only captures the semantic similarity of temporal progression between the generated and reference reports, but also enforces time point alignment. Furthermore, the evaluation is performed at the entity level, ensuring that temporal changes are aligned with the correct spatial entities. However, its sequential evaluation setting has not yet been compared with radiologist judgments, and the sequential dataset comprises a limited sample of only 10 patients, which restricts its effectiveness in assessing the quality of sequential time-point reports. These limitations indicate a promising direction for future improvement.

\begin{figure}[!t]
    \includegraphics[width=\columnwidth]{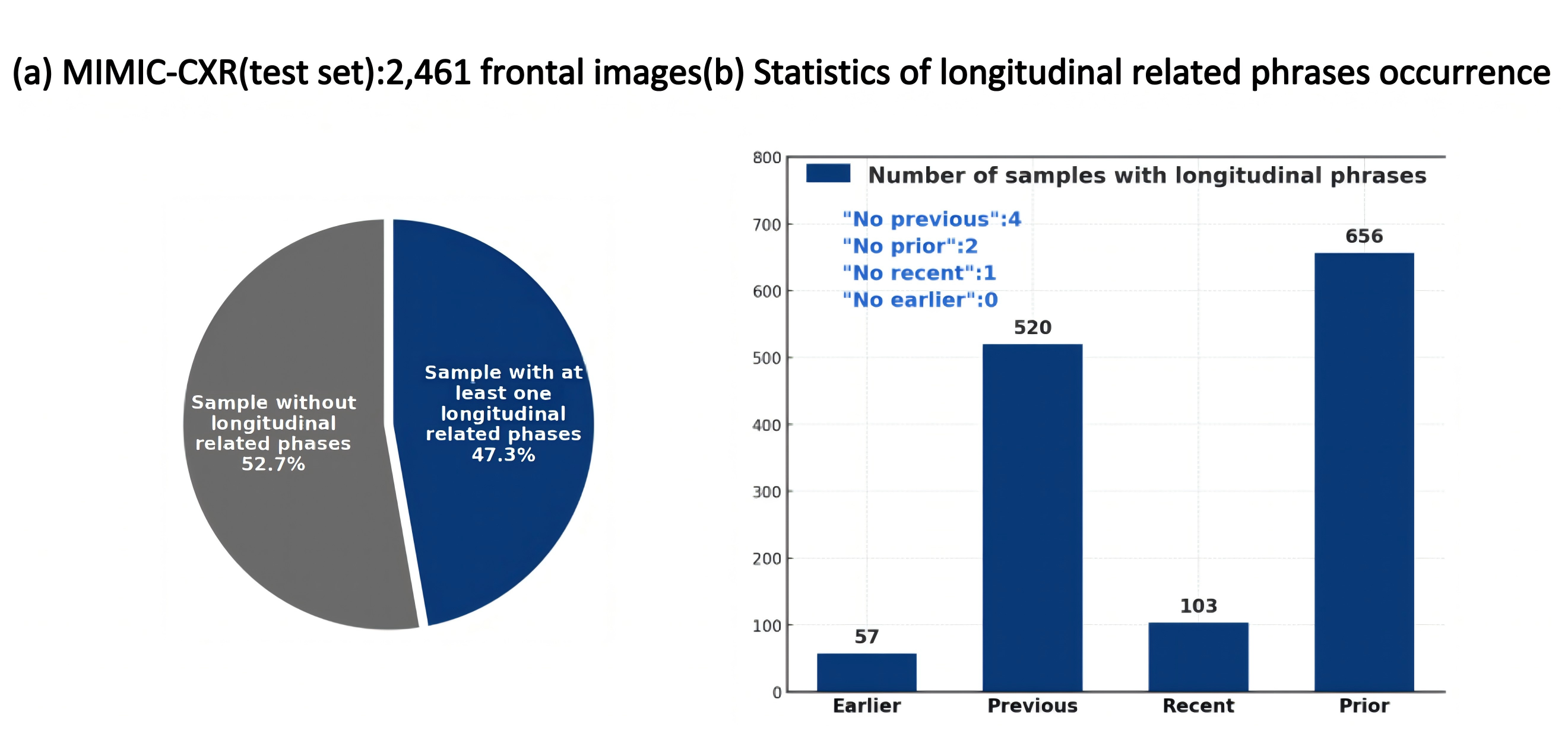}
    \caption{Distribution of longitudinally related phrases in the MIMIC-CXR test set.
(a) Proportion of frontal-view studies (2,461 total) containing at least one such phrase (47.3\%) vs. none (52.7\%).
(b) Frequency of individual phrases: Earlier (57), Previous (520), Recent (103), Prior (656), No previous (4), No prior (2), No recent (1), No earlier (0).}
    \label{fig:longitudinal phases}
\end{figure}

\section{Discussion}
\label{sec:Discussion}
While the preceding sections have reviewed representative LRRG architectures and evaluation protocols, this section provides a structured summary of reported results, presented in Tables~\ref{tb:NLG comparison MIMIC}, \ref{tb:CE comparison MIMIC}, \ref{tb:New CE metric MIMIC}, and~\ref{tb:Longitudinal dataset comparison}. The use of longitudinal data introduces multiple sources of variability, complicating the interpretation of performance scores in isolation. To improve clarity, we explicitly document these factors. Specifically:  

\begin{itemize}
    \item Table~\ref{tb:NLG comparison MIMIC} lists different methods' vision-encoder-supported input resolutions, the evaluated text sections generated by their text decoders, and the modalities used of the longitudinal data. 
    \item Table~\ref{tb:CE comparison MIMIC} differentiates models by their handling of CheXbert~\citep{smit2020chexbert} labels.  
    \item With respect to the dataset scale as mentioned in Section~\ref{sec:Datasets}. In Table~\ref{tb:Longitudinal dataset comparison}, we indicate methods that rely solely on longitudinal cases. 
\end{itemize}

By consolidating these details, we aim to provide a transparent reference to support consistent and informed evaluation of LRRG methods.  

In the remainder of this section, we examine the role of longitudinal data in the RRG task and highlight the key components that facilitate the transition from conventional RRG to LRRG methods. Our discussion is informed by ablation studies reported in the original papers on the MIMIC-CXR test set. The section is organized as follows:  

\begin{itemize}
    \item \textbf{Overall impact of longitudinal data:}~Evidence from case studies, together with ablation experiments comparing longitudinal and single-image settings, highlights the value of temporal context.
    \item \textbf{Longitudinal-specific architectures:}~Analysis of structural components that effectively exploit temporal information to guide future LRRG research.
\end{itemize}

\subsection{Overall impact of longitudinal data}
In this section, we demonstrate the rich temporal information contained within the MIMIC-CXR dataset~\citep{johnson2019mimic}. Furthermore, we implement a backbone model to verify whether incorporating longitudinal data can truly enhance model performance. To provide additional evidence, we also include ablation studies from existing LRRG methods comparing single-time-point and longitudinal settings. Specifically, we analyzed the findings sections of all frontal-image reports in the MIMIC-CXR test set (Fig.~\ref{fig:longitudinal phases}). We counted the phrases “prior'', “previous'', “recent'', and “earlier” as markers of longitudinal information. To avoid misclassification of statements such as “No previous study is available'', we also counted the phrases “No prior'', “No previous'', “No recent'', and “No earlier''. Among 2,461 reports, 1,164 (47.3\%) contained at least one such phrase without the negation term “No'', underscoring the central role of temporal references in clinical practice. However, models trained solely on single images cannot capture this longitudinal component, as demonstrated in the following case study.

We adopted \textit{BLIP-3}~\citep{xue2024xgen} as the backbone for this case study, following the longitudinal data configuration of \textit{Libra}~\citep{zhang2024libra}. Experiments used the frontal view of MIMIC-CXR, with the current image replicated as a proxy when no prior was available. In the single-image setting, prompts incorporated clinical context (Indication, Comparison, History, and Technique), whereas in the longitudinal setting, prior findings were additionally included. 

For model training in both single-image and longitudinal settings, we set the batch size to 2 per device, the learning rate to 0.001, and trained for 3 epochs. The LoRA configuration used a rank of 64 with $\alpha$ set to 128. We employed the AdamW optimizer, and the model was implemented on four NVIDIA GTX 3090 GPUs.

Model performance was evaluated using the Bleu-4 and METEOR metrics. On 2,461 frontal-view pairs from the MIMIC-CXR test set, the longitudinal BLIP-3 achieved a Bleu-4 score of 0.184 and a METEOR score of 0.428, slightly outperforming the single-image BLIP-3 (0.088 and 0.306, respectively).

Figure~\ref{fig:Case study} further compares the generated reports from single-image and longitudinal BLIP-3 with the reference report. The single-image BLIP-3 produced fewer correct findings and overlooked critical comparisons—for example, describing the cardiac silhouette as “unchanged” rather than the reference description of “continued enlarged''. In contrast, the longitudinal BLIP-3 effectively leveraged temporal context. Both the qualitative case study and quantitative results underscore that, without longitudinal information, models struggle to capture progression and comparative cues.

For more comprehensive evidence, we further summarize ablation results from different methods. As shown in Fig.~\ref{fig:visual_single_vs_longitudinal}, although the degree of improvement varies across methods, all longitudinal settings consistently surpass their single-image counterparts. This confirms that longitudinal modeling can effectively address this challenge and enable more accurate and clinically meaningful report generation.

Building on this evidence, we next aggregate the documented ablation studies' results from the existing LRRG methods to identify the architectural components that enable models to effectively leverage longitudinal features. 


\begin{table*}[htbp]
\centering
\caption{
Performance of Longitudinal Models on the MIMIC-CXR Test Set. Models are grouped by input type (image-only, report-only, image–report combined). Evaluation covers Findings and Impression sections using Bleu-1 to Bleu-4, METEOR, and ROUGE-L.
\ding{171} denotes models trained with frontal images only. Scores for the Impression section are shown in grey.}
\label{tb:NLG comparison MIMIC}
\resizebox{\textwidth}{!}{
    \begin{tabular}{llclcccccccc}
        \toprule
        \textbf{Scale of the longitudinal input} & \textbf{Model} & \textbf{Year} & \textbf{Eval. section} & \textbf{Input size} & \textbf{Bleu-1} & \textbf{Bleu-2} & \textbf{Bleu-3} & \textbf{Bleu-4} & \textbf{METEOR} & \textbf{ROUGE-L} \\ 
        \midrule
        \multirow{5}{*}{Image-Only Prior Studies} 
        & CCXRG~\citep{serra2023controllable} \ding{171} & 2023 & Findings & 512 & 0.486 & 0.366 & 0.295 & 0.246 & 0.216 & 0.423 \\
        & MedVersa~\citep{zhou2024medversa} & 2024 & Findings/{\color{gray}Impression} & 224 & - & - & - & 0.178/{\color{gray}0.137} & - & - \\
        & HERGen~\citep{wang2024hergen} \ding{171} & 2024 & Findings+Impression & 384 & 0.395 & 0.248 & 0.169 & 0.122 & 0.156 & 0.285 \\
        & ERRG~\citep{santiesteban2024enhancing} \ding{171}  & 2024 & Findings/{\color{gray}Impression} & 512 & - & - & - & 0.221/{\color{gray}0.214} & - & 0.170/{\color{gray}0.161} \\
        & Libra~\citep{zhang2024libra} \ding{171} & 2024 & Findings & 518 & 0.513 & - & - & 0.245 & 0.489 & 0.367 \\
        & TiBiX~\citep{sanjeev2024tibix} \ding{171} & 2024 & Findings+Impression & 512 & 0.324 & 0.234 & 0.185 & 0.157 & 0.162 & 0.331 \\
        & PriorRG~\citep{liu2025priorrg} & 2025 & Findings & 518 & 0.412 & 0.290 & 0.220 & 0.175 & 0.189 & 0.324 \\
        \midrule
        \multirow{2}{*}{Report-Only Prior Studies}
        & LDSSR~\citep{nicolson2024longitudinal} & 2024 & Findings & 384 & - & - & - & 0.079 & - & 0.262 \\
        & LLM-RG4~\citep{wang2025llm} & 2025 & Findings & 518 & 0.377 & - & - & 0.144 & - & 0.318 \\
        \midrule
        \multirow{13}{*}{Combined Image \& Report Prior Studies}
        & RECAP~\citep{hou2023recap} & 2023 & Findings & 224 & 0.429 & 0.267 & 0.177 & 0.125 & 0.168 & 0.288 \\
        & BioVil-T~\citep{bannur2023learning} \ding{171} & 2023 & Findings+Impression/{\color{gray}Impression} & 448 & - & 0.213/{\color{gray}0.159} & - & - & - & - \\
        & EVOKE~\citep{miao2024evoke} & 2024 & Findings & 384 & 0.408 & 0.271 & 0.197 & 0.151 & 0.171 & 0.313 \\
        & MAIRA-2~\citep{bannur2024maira} & 2024 & Findings & 518 & 0.460 & - & - & 0.231 & 0.417 & 0.384 \\
        & DDaTR~\citep{song2025ddatr} & 2025 & Findings & 224 & 0.401 & - & - & 0.113 & 0.162 & 0.272 \\
        & RADAR~\citep{hou2025radar} \ding{171} & 2025 & Findings & 384 & 0.509 & - & - & 0.262 & 0.450 & 0.397 \\
        & MLRG~\citep{liu2025enhanced} & 2025 & Findings/Findings+Impression & 518 & 0.411/{0.402} & 0.277/{0.270} & 0.204/{0.197} & 0.158/{0.152} & 0.176/{0.172} & 0.320/{0.327} \\
        & STREAM~\citep{yang2025spatio} & 2025 & Findings & 224 & 0.437 & 0.278 & 0.192 & 0.139 & 0.172 & 0.297 \\
        \bottomrule
    \end{tabular}
}
\end{table*}

\begin{table}[!t]
\centering
\caption{Performance on the MIMIC-CXR test set under different CheXbert label strategies. Grey scores treat uncertain labels as positive, “*” as negative, and unmarked as a separate class. \ding{171} denotes models trained with frontal images only. For RADAR, two methods are reported (distinguished by square brackets).}
\label{tb:CE comparison MIMIC}
\resizebox{\columnwidth}{!}{
\begin{tabular}{llccc}
\toprule
\textbf{Scale of the longitudinal input} & \textbf{Model} & \textbf{Precision$_{\text{macro/micro}}$} & \textbf{Recall$_{\text{macro/micro}}$} & \textbf{F1$_{\text{macro/micro}}$} \\
\midrule
\multirow{5}{*}{Image-Only Prior Studies}
& CCXRG~\citep{serra2023controllable} \ding{171} & {\color{gray}/0.597} & {\color{gray}/0.516} & {\color{gray}/0.553} \\
& HERGen~\citep{wang2024hergen} \ding{171} & 0.415/ & 0.301/ & 0.317/ \\
& Libra~\citep{zhang2024libra} \ding{171} & - & - & 0.404/0.559 \\
& TiBiX~\citep{sanjeev2024tibix} \ding{171} & 0.300/ & 0.224/ & 0.250/ \\
\midrule
\multirow{2}{*}{Report-Only Prior Studies}
& LDSSR~\citep{nicolson2024longitudinal} & 0.438*/ & 0.349*/ & 0.357*/ \\
& LLM-RG4~\citep{wang2025llm} & /{\color{gray}0.583} & {\color{gray}/0.593} & {\color{gray}/0.588} \\
\midrule
\multirow{13}{*}{Combined Image \& Report Prior Studies}
& RECAP~\citep{hou2023recap} & {\color{gray}0.389/} & {\color{gray}0.443}/ & {\color{gray}0.393/} \\
& BioVil-T~\citep{bannur2023learning} \ding{171} & - & -& 0.359/ \\
& EVOKE~\citep{miao2024evoke} & 0.400/0.538 & 0.350/0.465 & 0.354/0.499 \\
& MAIRA-2~\citep{bannur2024maira} & - & - & 0.416*/0.581* \\
& M4CXR~\citep{park2024m4cxr} & - & - & 0.420*/0.607* \\
& DDaTR~\citep{song2025ddatr} & {\color{gray}0.438/} & {\color{gray}0.465/}& {\color{gray}0.441/} \\
& RADAR~\citep{hou2025radar} \ding{171} & - & - & 0.460[{\color{gray}0.497}]/0.653[{\color{gray}0.674}] \\
& MLRG~\citep{liu2025enhanced} & 0.440/0.549 & 0.354/0.468 & 0.364/0.505 \\
& STREAM~\citep{yang2025spatio} & {\color{gray}/0.515} & {\color{gray}/0.447} & {\color{gray}/0.478} \\
& PriorRG~\citep{liu2025priorrg} & 0.429/0.541 & 0.368/0.485 & 0.376/0.511 \\
\bottomrule
\end{tabular}
}
\end{table}

\subsection{Importance of longitudinal-specific architectures}
Based on the methods reviewed in Section~\ref{sec:Core Challenges and Approaches in Longitudinal Report Generation}, two primary strategies directly address longitudinal features: feature alignment and feature fusion, while additional enhancement modules also contribute substantially. To assess their impact, within-method ablation results reported under controlled conditions were extracted from the original publications to isolate the contribution of each component, as illustrated in Fig.~\ref{fig:Ablation study data}.

\begin{table}[htbp]
\centering
\caption{Performance of Other Clinical Scores on the MIMIC-CXR Test Set. \ding{171} denotes models trained with frontal images only. Metrics where lower values indicate better performance are marked with “$\downarrow$''. }
\label{tb:New CE metric MIMIC}
\resizebox{\columnwidth}{!}{
\begin{tabular}{lcccc}
\toprule
\textbf{Model} & \textbf{Chexbert vector} & \textbf{RG (F1)} & $\mathbf{RG}_{\mathbf{ER}}$ & \textbf{RadCliQ $\downarrow$} \\
\midrule
MedVersa~\citep{zhou2024medversa} & 46.4 & 0.280 & - & 2.71 \\
LDSSR~\citep{nicolson2024longitudinal} & - & - & 27.2 & - \\
BioVil-T~\citep{bannur2023learning} \ding{171} & 35.9 & - & - & - \\
EVOKE~\citep{miao2024evoke} & - & 0.278 & - & - \\
MAIRA-2~\citep{bannur2024maira} & 50.7 & 0.346 & 39.6 & 2.64 \\
Libra~\citep{zhang2024libra} \ding{171} & 46.9 & 0.329 & 37.3 & 2.72 \\
DDaTR~\citep{song2025ddatr} & - & 0.198 & - & 1.14 \\
RADAR~\citep{hou2025radar} \ding{171} & - & 0.346 & 39.3 & 2.61 \\
MLRG~\citep{liu2025enhanced} & - & 0.291 & - & - \\
STREAM~\citep{yang2025spatio} & 40.5 & 0.223 & - & - \\
\bottomrule
\end{tabular}
}
\end{table}

\subsubsection{Contrastive Learning as a Key Module in the LRRG methods}
\begin{figure}[!t]
    \centering
    \includegraphics[width=\linewidth]{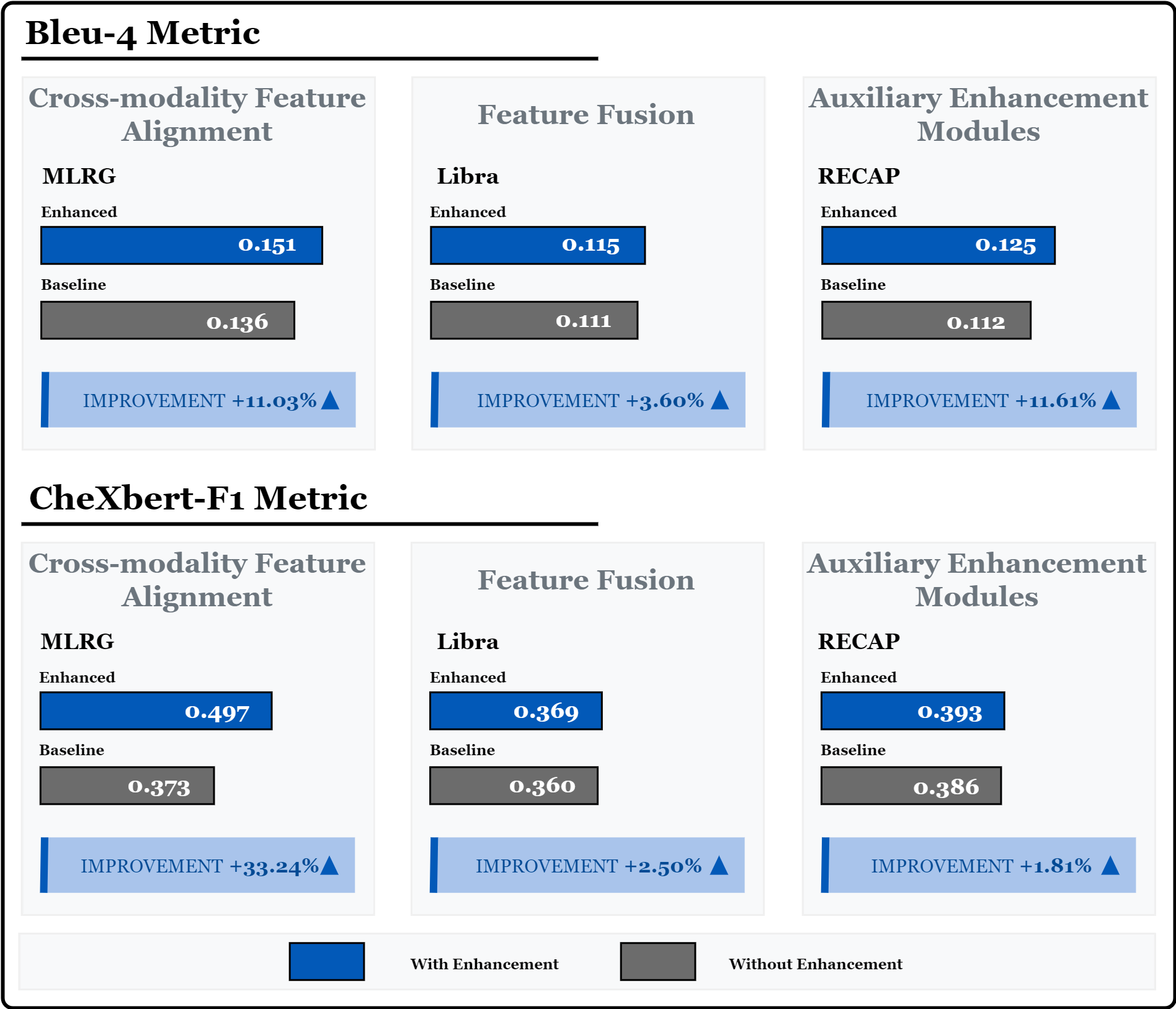}
    \caption{Ablation results of representative LRRG methods (\textit{MLRG}~\citep{liu2025enhanced}, \textit{Libra}~\citep{zhang2024libra}, and \textit{RECAP}~\citep{hou2023recap}). The figure shows consistent performance drops when key components are removed, highlighting the contributions of contrastive learning, feature fusion, and auxiliary modules across different architectural designs.}
    \label{fig:Ablation study data}
\end{figure}

Contrastive learning constitutes a core mechanism for aligning features across modalities. The ablation of \textit{MLRG}~\citep{liu2025enhanced} analysis in Fig.~\ref{fig:Ablation study data} substantiates the importance of contrastive learning, showing that removing the alignment objective of \textit{MLRG} decreases Bleu-4 from 0.151 to 0.136 and micro-CheXbert F1 from 0.497 to 0.373. \textit{MLRG} has demonstrated the effectiveness of batch-level cross-modal alignment, while \textit{HC-LLM}~\citep{liu2025hc} further reveals that intra-patient alignment across different time points can also enhance model performance, as evidenced by its ablation results in \textit{Table II} of its original paper~\citep{liu2025hc}. When contrastive objectives are enabled, the model achieves a Bleu-4 score of 0.128 and a CheXbert F1 score of 0.357, whereas removing intra-patient alignment decreases these metrics to 0.118 and 0.311, respectively. In the LRRG task, temporal variations in imaging must be coherently captured in text, yet radiology reports often contain abundant details beyond the actual cross-time changes, which may introduce redundancy or noise. By enforcing cross-time and cross-modality contrastive alignment at both batch and patient levels between fused longitudinal visual features and their corresponding reports, the model learns to identify positive pairs based on temporal variations. This alignment enhances its capacity to capture clinically meaningful disease progression and ensures consistency between imaging and textual representations.

Overall, the evidence indicates that contrastive learning is foundational for capturing consistency across time and modalities. At the same time, performance still depends on how aligned features are ultimately fused, underscoring fusion as the next critical design factor.

\subsubsection{Feature fusion as a Key Module in the LRRG methods}
Feature fusion plays a central role in the LRRG task, as it enables models to effectively integrate information across temporally distributed imaging studies. By emphasizing clinically relevant temporal signals and attenuating redundant or noisy features, fusion mechanisms support the extraction of progression cues critical to accurate report generation. While only using the frontal view of the Chest X-ray images, \textit{Libra}~\citep{zhang2024libra} shows the strong performance of its Temporal Fusion Module (TFM), which leverages attention mechanisms to integrate cross-temporal visual features. The ablation analysis in Fig.~\ref{fig:Ablation study data} proves TFM contribution, showing that removing this module reduces Bleu-4 from 0.115 to 0.111 and CheXbert F1 from 0.369 to 0.360. A consistent trend appears in \textit{DDaTR}~\citep{song2025ddatr}, which designs a dedicated cross-time fusion module (DDAM). The ablation results in the original paper of~\citep{song2025ddatr} confirm the effectiveness of the proposed module, showing that removing DDAM leads to marked performance drops, with Bleu-4 declining from 0.113 to 0.100 and CheXbert F1 from 0.441 to 0.402. These findings underscore the importance of fusion mechanisms in the LRRG methods, where models must prioritize spatial-temporal regions indicative of disease progression. Without effective fusion, extraneous temporal features can dilute critical signals and mislead clinical interpretation.

\subsubsection{The Augmentative Role of the Auxiliary Enhancement Module in the LRRG methods}

This subsection investigates the impact of auxiliary enhancement modules on model performance. As described in Section~\ref{sec:Core Challenges and Approaches in Longitudinal Report Generation}, \textit{RECAP}~\citep{hou2023recap} incorporates two auxiliary enhancement modules. These modules classify temporal changes across studies and construct a relational graph based on temporal and spatial entities. To evaluate their contributions, ablation studies were performed to examine the impact of each module on overall performance, as reported in Fig.~\ref{fig:Ablation study data} and \textit{Table IV} within its article~\citep{hou2023recap}.  Removing the temporal progression classification module decreases Bleu-4 from 0.125 to 0.112 and CheXbert F1 from 0.393 to 0.386. Eliminating the relational graph module results in similar declines (Bleu-4: 0.119; CheXbert F1: 0.391). When both modules are removed, performance declines further, with Bleu-4 reduced to 0.113 and CheXbert F1 to 0.296. These results demonstrate that incorporating auxiliary signals effectively enhances longitudinal radiology report generation by explicitly guiding the model to recognize temporal progression and leveraging retrieval-based augmentation to capture entity-level relations and their temporal evolution. Collectively, these signals enable more accurate characterization of disease dynamics and facilitate the generation of reports that more closely align with clinical reasoning.

\begin{table}[!htbp]
\centering
\caption{Performance of Different Methods on the Longitudinal MIMIC-CXR Test Set (Cases with Longitudinal Data Only).}
\label{tb:Longitudinal dataset comparison}
\resizebox{\columnwidth}{!}{
\begin{tabular}{lcccc}
\toprule
\textbf{Model} & \textbf{Eval. section} &\textbf{Bleu-4} & \textbf{ROUGE-L} & \textbf{F1$_{\text{macro/micro}}$} \\
\midrule
HC-LLM~\citep{liu2025hc} & Findings+Impression & 0.142 & 0.284 & /0.498 \\
ULCX~\citep{zhu2023utilizing} & Findings &0.099 & 0.271 & /0.480 \\
DDaTR~\citep{song2025ddatr} & Findings & 0.105 & 0.266 & /0.527 \\
HERGen~\citep{wang2024hergen} & Findings+Impression & 0.117 & 0.282 & 0.317/ \\
Diff-RRG~\citep{yun2025diff} & Findings+Impression & 0.120 & 0.276 & /0.474 \\
\bottomrule
\end{tabular}
}
\end{table}

\begin{figure}[!t]
    \centering
    \includegraphics[width=0.8\columnwidth]{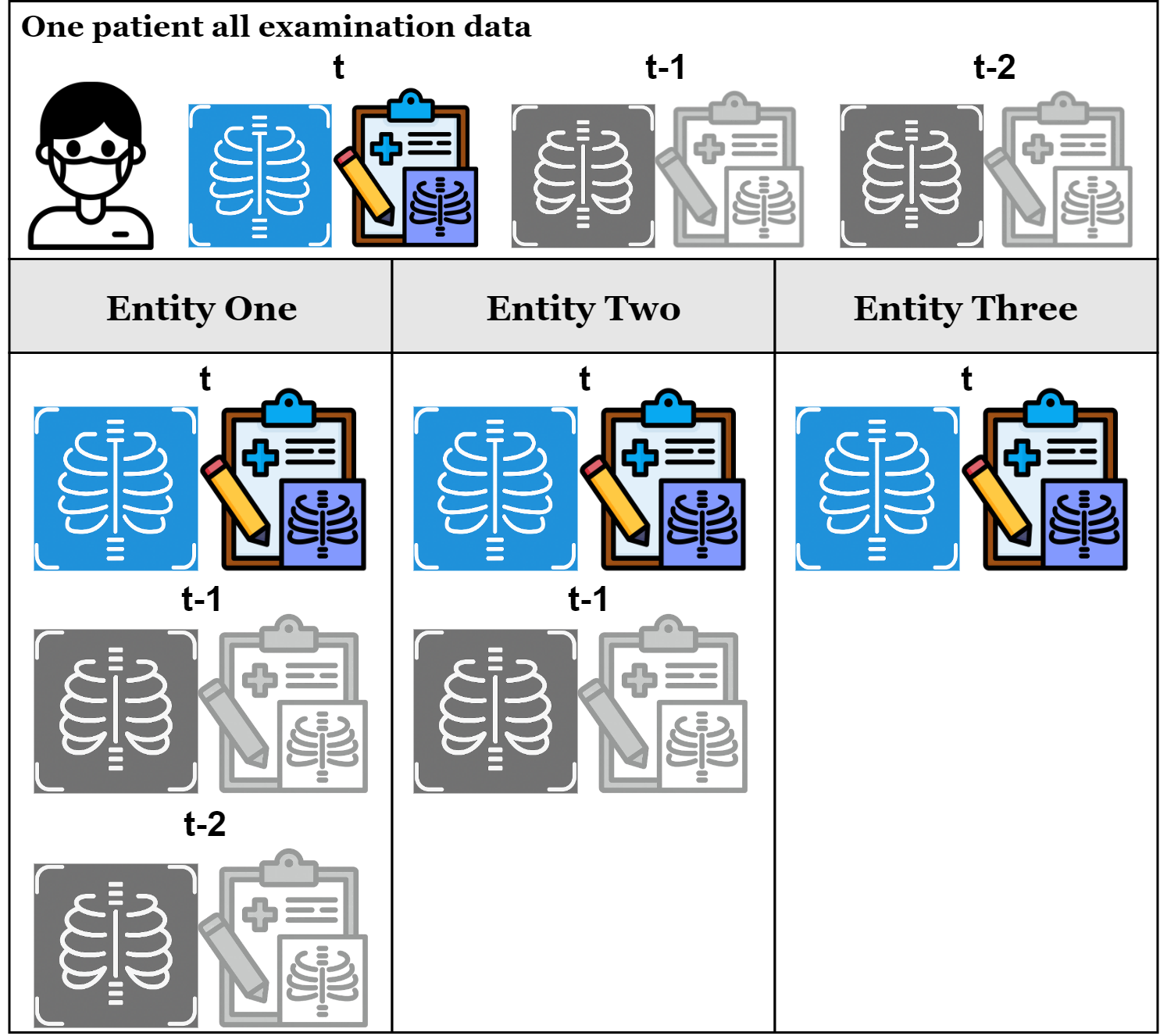}
    \caption{Illustration of longitudinal data augmentation by intra-patient resampling. The top row shows all available examinations of one patient at different time points (t, t-1, t-2). To enlarge the training set, multiple subsets are created by pairing the current study with its prior examinations. The lower panel illustrates three such augmented instances (Entity One–Three), where each instance reuses part of the same patient’s history to form a new training sample. }
    \label{fig:reuse image}
\end{figure}

\section{Current limitations and future improvements}
\label{sec:Current limitations and future improvements}
This section consolidates our analysis of LRRG by examining both its present limitations and avenues for future improvement. Identifying these challenges is crucial not only for understanding the boundaries of existing methods but also for charting research directions that can better leverage longitudinal information for clinically faithful report generation.

\subsection{Current limitations}

Despite recent progress, several limitations remain in current LRRG methods. Specifically, we identify five major limitations. 
\subsubsection{Insufficient availability of longitudinal data}Most existing LRRG studies rely on subsets of MIMIC-CXR, without a dedicated longitudinal dataset. To enlarge the longitudinal cases within the dataset, many methods augment training by reusing multiple studies from the same patient, shown in Fig.~\ref{fig:reuse image}, which increases sample size but reduces patient diversity and biases models toward short-term changes. The lack of comprehensive longitudinal datasets thus remains a critical barrier to LRRG advancement.

\subsubsection{Limited attention to temporal gaps between studies}Among LRRG methods, only a few explicitly consider time gaps between input studies, despite their importance for modeling temporal progression. Current sampling strategies often fix the earliest scan and randomly select subsequent ones before ordering them chronologically, a process that tends to overlook the continuity of disease progression. For recurrent conditions such as pleural effusion or edema, omitting recovery intervals obscures remission–relapse patterns, forcing comparisons against ill-suited priors and ultimately degrading progression assessment and report quality.

\subsubsection{Lack of strategies for Redundancy Mitigation}The use of longitudinal data inevitably increases the token count; however, not all prior images or reports are equally relevant for longitudinal comparison. Incorporating the entire history of studies introduces substantial redundancy and may even mislead the model, potentially degrading performance. The truly critical information often resides at specific time points where features indicative of disease progression are present.

\subsubsection{Inadequate consideration of image registration across time}
\begin{figure}[!t]
    \centering
    \includegraphics[width=0.7\columnwidth]{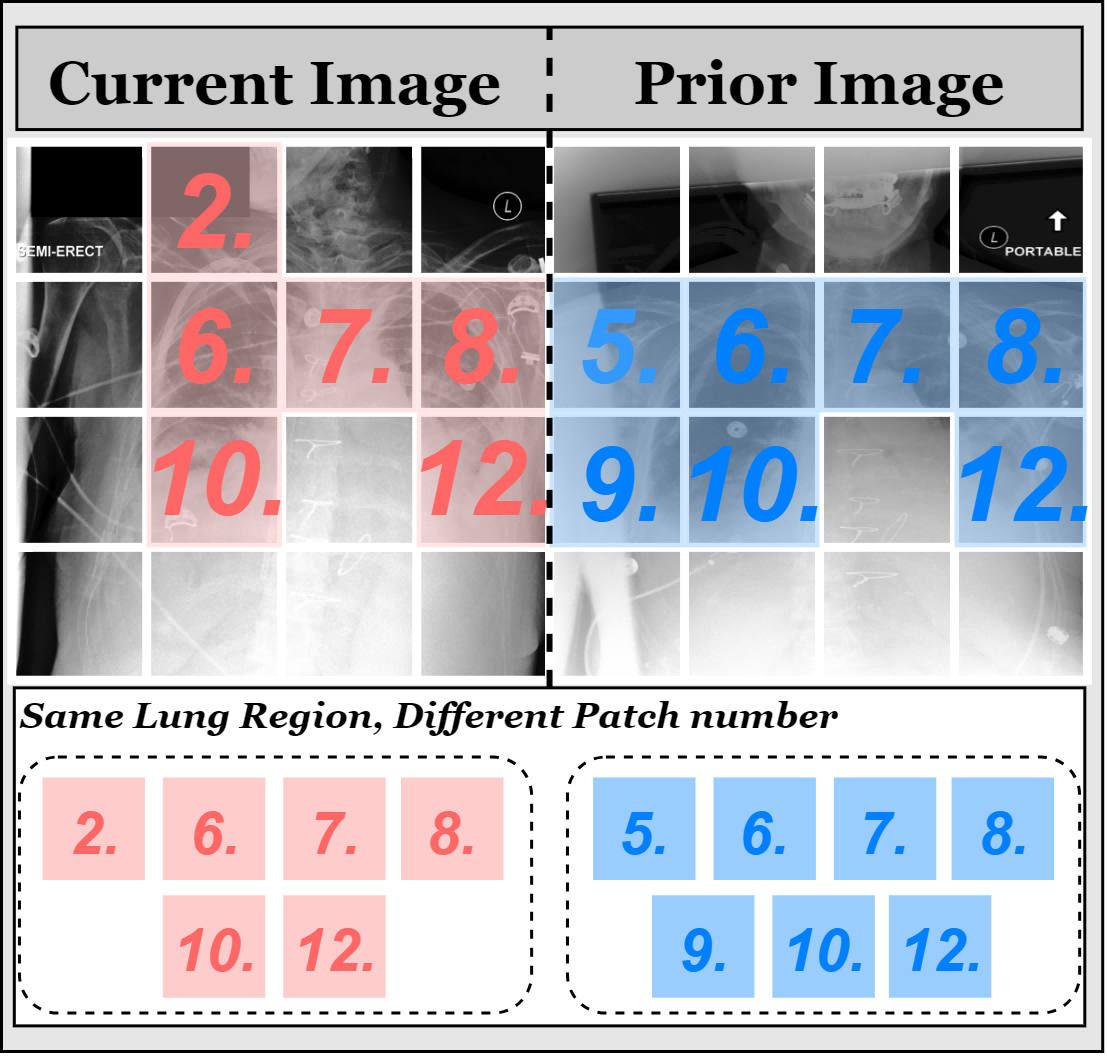}
    \caption{Illustration of positional inconsistencies in lung regions of the same patient across time. The red patches correspond to spatial divisions of the current Chest X-ray image, while the blue patches represent the corresponding prior image.}
    \label{fig:Position difference between patches}
\end{figure}

Many methods also neglect image registration across different time points. Variations in patient positioning, posture, or acquisition settings can cause identical anatomical structures to appear in different spatial locations across studies. As illustrated in Fig.~\ref{fig:Position difference between patches}, even within the lung region of a single patient, corresponding findings may fall into different patch locations. Without proper registration, this spatial misalignment can mislead the model and undermine its ability to capture genuine longitudinal changes.

\subsubsection{Lack of longitudinal-specific evaluation metrics and dataset}Many existing metrics mainly focus on matching clinical entities between the generated and reference reports, while overlooking the overlap of temporal progression entities. Moreover, current longitudinal-specific metrics lack large-scale and radiologist-validated evaluation data, posing challenges to accurately assessing a model’s ability to capture temporal progression information.

\subsection{Future improvements}
Having outlined the current limitations of LRRG methods, we next turn to potential avenues for progress. The following sections summarize prospective directions aimed at addressing these limitations and at broadening the scope of longitudinal modeling. 

\noindent\textbf{Leveraging Model-Generated Longitudinal Data:}~A promising future direction to address the scarcity of longitudinal cases is leveraging model-generated data as auxiliary input. \textit{LDSSR}~\citep{nicolson2024longitudinal} demonstrates that a report generated from a previous visit can be reused as pseudo-longitudinal context. Moreover, emerging generative paradigms—such as diffusion and flow-based models—have the potential to simulate disease evolution over time, thereby supporting data augmentation and benchmarking under controlled temporal trajectories. These advances suggest that model-generated longitudinal data, spanning multiple modalities, could serve as a valuable complement to real-world datasets.

\noindent\textbf{Time-Aware Longitudinal Modeling:}~Existing approaches often underutilize time-gap information. Future work may leverage temporal intervals to dynamically fuse prior studies, since not all historical features equally contribute to the current diagnosis. Moreover, aligning time gaps with patient-specific context, for example, a short interval following surgery may indicate substantial changes in disease status. By aligning temporal intervals with patient-specific clinical context (e.g., indications or other auxiliary records), models may achieve more precise and clinically meaningful longitudinal reasoning.

\noindent\textbf{Towards Redundancy-Aware Longitudinal Modeling:}~With the advent of mature token pruning techniques~\citep{vasu2025fastvlm, liu2025nvila}, models can discard redundant visual tokens to improve efficiency, while classification-based modules such as \textit{HC-LLM}~\citep{liu2025hc} can further filter temporal features, enabling models to focus on the most informative signals.

\noindent\textbf{Towards Spatially-Aligned Longitudinal Modeling:}~To effectively address the regression challenge posed by images from different time points, future work could incorporate spatial registration modules or learnable alignment strategies to normalize anatomical structures and ensure that temporal comparisons are conducted on consistent regions.

\noindent\textbf{Advancing Evaluation Frameworks for LRRG:}~To better assess the temporal reasoning ability of LRRG models, future research could focus on establishing large-scale datasets evaluated by radiologists, enabling more reliable validation of whether longitudinal-specific metrics align with expert clinical judgments. Moreover, due to the current absence of robust longitudinal-specific metrics, future work could also explore more comprehensive evaluation designs.

\noindent\textbf{Toward Disease Progression Prediction Based on Longitudinal Modeling:}~A promising future direction is to move beyond retrospective description and leverage longitudinal data for prospective prediction. By modeling disease trajectories, LRRG systems could provide clinicians with forecasts of a patient’s future condition, thereby assisting treatment planning and early intervention. Such models could also generate automatic alerts or risk indicators immediately after an examination, enriching radiology reports with actionable warnings.

\noindent\textbf{Toward Multimodal Generalization Beyond Chest X-rays:}~Current research is largely limited to Chest X-rays, while longitudinal report generation for other modalities, such as MRI and CT, remains underexplored. This focus reflects the current data availability rather than a methodological constraint. The modeling principles surveyed in this work are inherently modality-agnostic and could be adapted to other imaging domains mentioned above. Extending LRRG beyond Chest X-rays would introduce new challenges related to higher spatial and temporal granularity, three-dimensional anatomical structures, and diverse reporting styles, but it also represents an important step toward a unified framework for longitudinal reasoning across imaging modalities.

\section{Conclusion}
\label{sec:Conclusion}
In this paper, we surveyed recent advances in the LRRG task, an emerging task that extends conventional RRG by incorporating prior studies to enhance clinical fidelity. We reviewed dataset construction strategies, model architectures for generating the radiology report and integrating longitudinal information, as well as evaluation protocols, including longitudinal-specific metrics. We further compile performances as well as variables of different LRRG methods and analyze ablation studies of representative models to underscore the critical role of longitudinal data and structural design choices in improving model performance. Based on these observations, we identified five major limitations along with corresponding potential solutions, as well as prospective directions for leveraging longitudinal data. Overall, this survey provides a comprehensive synthesis of existing work, underscores the key challenges of LRRG, and outlines promising research directions to guide future progress.

\appendix




\bibliography{cas-refs}
\end{document}